\documentclass[10pt,journal,compsoc]{IEEEtran}

\usepackage{amsmath}

\usepackage{amssymb}
\usepackage{mathrsfs}

\usepackage{bm}

\ifCLASSOPTIONcompsoc
  \usepackage[nocompress]{cite}
\else
  \usepackage{cite}
\fi

\ifCLASSINFOpdf
  \usepackage[pdftex]{graphicx}
  \DeclareGraphicsExtensions{.pdf}
\else

\fi

\usepackage{graphicx}

\begin{document}

\title{Robust Ellipsoid Fitting Using \\ Axial Distance and Combination}

\author{Min Han, Jiangming Kan, Gongping Yang, and Xinghui Li
\IEEEcompsocitemizethanks{
  \IEEEcompsocthanksitem M. Han and X. Li are with Tsinghua Shenzhen International Graduate School, Tsinghua University, Shenzhen, China. E-mail: hanm21@mails.tsinghua.edu.cn, li.xinghui@sz.tsinghua.edu.cn
\IEEEcompsocthanksitem J. Kan is with the School of Technology, Beijing Forestry University, Beijing, China. E-mail: kanjm@bjfu.edu.cn

\IEEEcompsocthanksitem G. Yang is with the School of Software, Shandong University, Jinan, China. E-mail: gpyang@sdu.edu.cn

}
\thanks{Manuscript received XX XX. XX; revised XX XX, XX.\\
Digital Object Identifier no. XX.XX/XX.XX}}

\markboth{IEEE TRANSACTIONS ON XX, VOL. xx, NO. xx, xx xx}%
{Shell \MakeLowercase{\textit{et al.}}: Bare Demo of IEEEtran.cls for Computer Society Journals}

\IEEEtitleabstractindextext{%
\begin{abstract}
  In random sample consensus (RANSAC), the problem of ellipsoid fitting can be formulated as a problem of minimization of point-to-model distance, which is realized by maximizing model score. Hence, the performance of ellipsoid fitting is affected by distance metric. In this paper, we proposed a novel distance metric called the axial distance, which is converted from the algebraic distance by introducing a scaling factor to solve nongeometric problems of the algebraic distance. There is complementarity between the axial distance and Sampson distance because their combination is a stricter metric when calculating the model score of sample consensus and the weight of the weighted least squares (WLS) fitting. Subsequently, a novel sample-consensus-based ellipsoid fitting method is proposed by using the combination between the axial distance and Sampson distance (CAS). We compare the proposed method with several representative fitting methods through experiments on synthetic and real datasets. The results show that the proposed method has a higher robustness against outliers, consistently high accuracy, and a speed close to that of the method based on sample consensus.
\end{abstract}

}

\maketitle

\IEEEdisplaynontitleabstractindextext
\IEEEpeerreviewmaketitle
\IEEEraisesectionheading{\section{Introduction}\label{sec:introduction}}

\IEEEPARstart{A}{S} a type of technology for pattern recognition, robust estimation of a quadratic surface has a wide range of applications. Compared to a sphere, an ellipsoid has more advantages in representing objects in the real world because it has three semiaxes \cite{A1}. For instance, ellipsoids have been applied to fit the limbs of humans for multiview gait recognition \cite{A2}. An ellipsoid is a good geometric primitive for tracking human heads \cite{A3,A4}. The calibrations of a magnetic compass and a camera are realized using ellipsoid fitting \cite{A5,A6}. Ellipsoid fitting algorithms can be applied to ellipse fitting after a minor modification when considering an ellipse as a low-dimensional ellipsoid. Ellipse fitting can address a lot of problems in the real world, e.g., gaze tracking \cite{A35}, shape control of silicon single crystal growth \cite{A36}, detection of forgings inclination \cite{A37}, measurements of complex voltage ratio \cite{A38}, quadrature imbalance compensation of microwave radar sensing \cite{A39}. Therefore, it is significant to study the ellipsoid fitting.

Depending on whether or not it is necessary to reduce the effect of outliers, the ellipsoid fitting methods can be divided into direct fitting methods \cite{A7,A8,A9,A10} and robust clustering methods \cite{A11,A12,A13,A14,A15,A16}.

As a representative of clustering methods, the sample-consensus-based method (SC-based method) often achieves a high robustness against outliers \cite{A17,A18}. RANSAC is an original version of the sample consensus and has been used to estimate 2-D ellipses and 3-D ellipsoids \cite{A11,A13,A14,A15,A16}. In each iteration, RANSAC mainly performs sampling, model fitting, model evaluation using model score, and updates of the best model according to model score. The best model with the highest model score is obtained after the required number of iterations. FLO-RANSAC (FLO) adds a local optimization step using WLS fitting into RANSAC to improve the fitting accuracy and speed \cite{A19}. Graph-cut-RANSAC (GC) introduces an efficient local optimization step realized using the graph-cut algorithm \cite{A20}. Even so, the SC-based methods have the risk of low accuracy owing to the random sampling and increased noise level. Additionally, the required number of iterations for the sample consensus is large for an ellipsoid model determined by nine points at least. Therefore, SC-based ellipsoid fitting methods are slower than direct fitting methods.

Unlike the SC-based methods, the direct fitting methods put an entire point set into the fitting pool to minimize the sum of squared distances between all points and a model. The fitting result may not be an ellipsoid or ellipse. A semi-definite program (SDP) is introduced to guarantee the ellipse solution \cite{A21}. An alternating direction method of multipliers (ADMM) is applied to simplify the computation of ellipsoid-specific fitting \cite{A9}. The direct fitting methods are often employed as an efficient estimation. Computing speed has increased with the development of computer technology. The fitting accuracy and robustness against outliers have become technical bottlenecks for some applications. For instance, we mainly consider accuracy rather than speed in camera calibration based on ellipsoid fitting \cite{A6}. Therefore, this paper focuses on fitting accuracy and robustness against outliers.

In terms of the geometric property of the distance metric, ellipsoid fitting methods can be classified into geometric distance-based methods and nongeometric distance-based methods. The common geometric distance includes the orthogonal distance \cite{A7,A22,A23,A24} and the Sampson distance \cite{A25,A26}. The nongeometric distance refers to the algebraic distance \cite{A8,A9,A27,A28}. The design of the distance metric directly affects the performance of ellipsoid fitting because it constructs the objective function of fitting. For example, the polar-n-direction distance, which is more accurate compared to the Sampson distance, is proposed to improve the fitting accuracy of a 2-D ellipse \cite{A29}.

The algebraic distance is defined as a functional value of an algebraic equation of the measured model at a measured point \cite{A8}. The isosurfaces (called isolines in 2-D space) of algebraic distance have the same shape as the measured model. The algebraic distance is nongeometric. However, the non-geometric property has limitations. The first one is fitting instability, and the fitting varies with coordinate transformation \cite{A7}. The second one is that the algebraic distance may result in a large fitting residual of the geometric distance. The third one is the lack of a physical interpretation, which results in some difficulties for the application (e.g., the setting of the threshold of algebraic distance).

The orthogonal distance is the shortest distance from a point to a model and is equal to the distance between the measured point and a foot point, which can be determined by two conditions: 1) the foot point is on the measured model, 2) the foot point is perpendicular to the measured model. The orthogonal distance is computationally expensive and excessively time-consuming. As a first-order approximation of the orthogonal distance, the Sampson distance is computationally simple and is calculated based on the orthogonality to contour \cite{A25,A29}.

SC-based methods have higher robustness against outliers than direct fitting methods. However, the SC-based methods have the risk of low accuracy owing to the random sampling and increased noise level. To achieve consistently high accuracy and higher robustness, our contributions are threefold:

1) We propose a novel geometric axial distance, which is converted from the algebraic distance by introducing a notion of scaling factor. The axial distance solves the nongeometric problems of the algebraic distance. The scaling factor has good physical interpretability for the modeling of an ellipsoid family.

2) The combination between the axial distance and Sampson distance (CAS) is proposed to form a strict metric and provides more constraint than a single distance.

3) We propose an SC-based ellipsoid fitting method where CAS is used as the point-to-model distance. The model score and weight in the WLS fitting are calculated by using CAS when considering the strict metric. Finally, the accuracy and robustness of the proposed method are improved without reducing the speed of sample consensus. The code and test datasets are available at (https://github.com/Mindyspm/CAS).

The remainder of this paper is organized as follows. Section II introduces preliminaries. Section III introduces the scaling factor, axial distance, and SC-based ellipsoid fitting using CAS. Section IV evaluates the proposed method by a vertical comparison with the direct fitting methods and a parallel comparison with the SC-based methods on synthetic and real datasets. Section V discusses limitations of the proposed method. Section VI concludes the paper and presents directions for future work.


\section{Preliminaries}

\subsection{Sample Consensus}
Linear least-square (LLS) \cite{A30} and WLS fittings are used to generate models in the model fitting stage. The model score is used to evaluate the quality of the models. Based on the model score, the sample consensus identifies the best model among the models generated by model fitting \cite{A11}. The cost function of model fitting and model score are constructed by point-to-model distance. Therefore, the design of the distance metric is extremely crucial to fitting results.

Suppose that $C$ denotes the total number of points in an input point set $\Omega  \subseteq {\mathbb{R}^3}$, $n$ denotes the size of a sample, and ${{\mathbf{x}}_h} = {[{x_1},{x_2},{x_3},1]^T} \in {\mathbb{R}^4}$ denotes the homogenous coordinates of an arbitrary point $p$. The general flow of finding the best model is introduced as follows.
First, an $n$-sized sample is drawn from the point set $\Omega$. Subsequently, a model $M$ is generated using LLS fitting with the $n$-sized sample. The sample consensus uses the model score to evaluate the model $M$. The point energy $\varphi$ denotes the contribution of a point to the model score. The RANSAC uses a top-hat fitness function to calculate the point energy. Corresponding to the model $M$, the point energy ${\varphi _{\left\{ {0,1} \right\}}}$ of the point $p$ with the top-hat fitness is
\begin{eqnarray}\label{eq1}
{\varphi _{\left\{{0,1} \right\}}}\!\left({p,M,\varepsilon } \right) = \left\{ \begin{gathered}
  1 - {\text{inlier       \ \ \ if }}d\left( {p,M} \right) < \varepsilon  \hfill \\
  0 - {\text{outlier    \ if }}d\left( {p,M} \right) \geqslant \varepsilon  \hfill \\ 
\end{gathered}  \right.,
\end{eqnarray}
\noindent where $d\left(  \cdot  \right)$ denotes the operator of point-to-model distance; $\varepsilon$ is a predefined distance threshold. This calculation way of point energy is too rigid. The references \cite{A20} and \cite{A31} suggest applying a Gaussian-kernel function to normalize the point energy into the range of 0–1 to evaluate the contribution of each point to the model score more accurately. The point energy ${\varphi _G}$ with the Gaussian kernel function is
\begin{eqnarray}\label{eq2}
{\varphi _G}\!\left( {p,M,\varepsilon } \right) = \exp \left( { - \frac{{d{{\left( {p,M} \right)}^2}}}{{2{\varepsilon ^2}}}} \right).
\end{eqnarray}

After the energy ${\varphi _G}$ of each point is calculated, the model score $\mathcal{S}$ is expressed as
\begin{eqnarray}\label{eq3}
\mathcal{S}\!\left( {M,\varepsilon } \right) = \sum\limits_{p \in \Omega } {{\varphi _G}\! \left( {p,M,\varepsilon } \right)} .
\end{eqnarray}

The best model is updated based on the model score. Thereafter, the best inliers are updated according to the point classification in Eq.~\eqref{eq1}. The final best model will be determined by iteration of the above process. The required number of iterations $J(v,\mu ,n)$ is \cite{A11}:
\begin{eqnarray}\label{eq4}
J(v,\mu ,n) = log\left( {1 - \mu } \right)/log\left( {1 - {v^n}} \right),
\end{eqnarray}

\noindent where $v$ denotes the ratio of the current best inliers corresponding to the current best model and is variable with the iteration, $\mu $ denotes the confidence level, and $n$ denotes the number of points in a sample. The guarantee of an ellipsoid solution is a challenge for the direct fitting methods \cite{A21}. However, SC-based methods can easily guarantee an ellipsoid solution using model validation.

\subsection{Complementarity of Distances}

The orthogonal distance is calculated using the orthogonality between the measured point and measured model. The Sampson distance is the first-order approximation of the orthogonal distance and is calculated based on the orthogonality of the measured point to the contour of the measured model \cite{A25,A29}. Therefore, the measurement ways of the orthogonal distance and Sampson distance are similar. The small difference of metrics means they may be non-complementary.

\begin{figure}[!t]
\centering
\includegraphics[width=0.8 \linewidth]{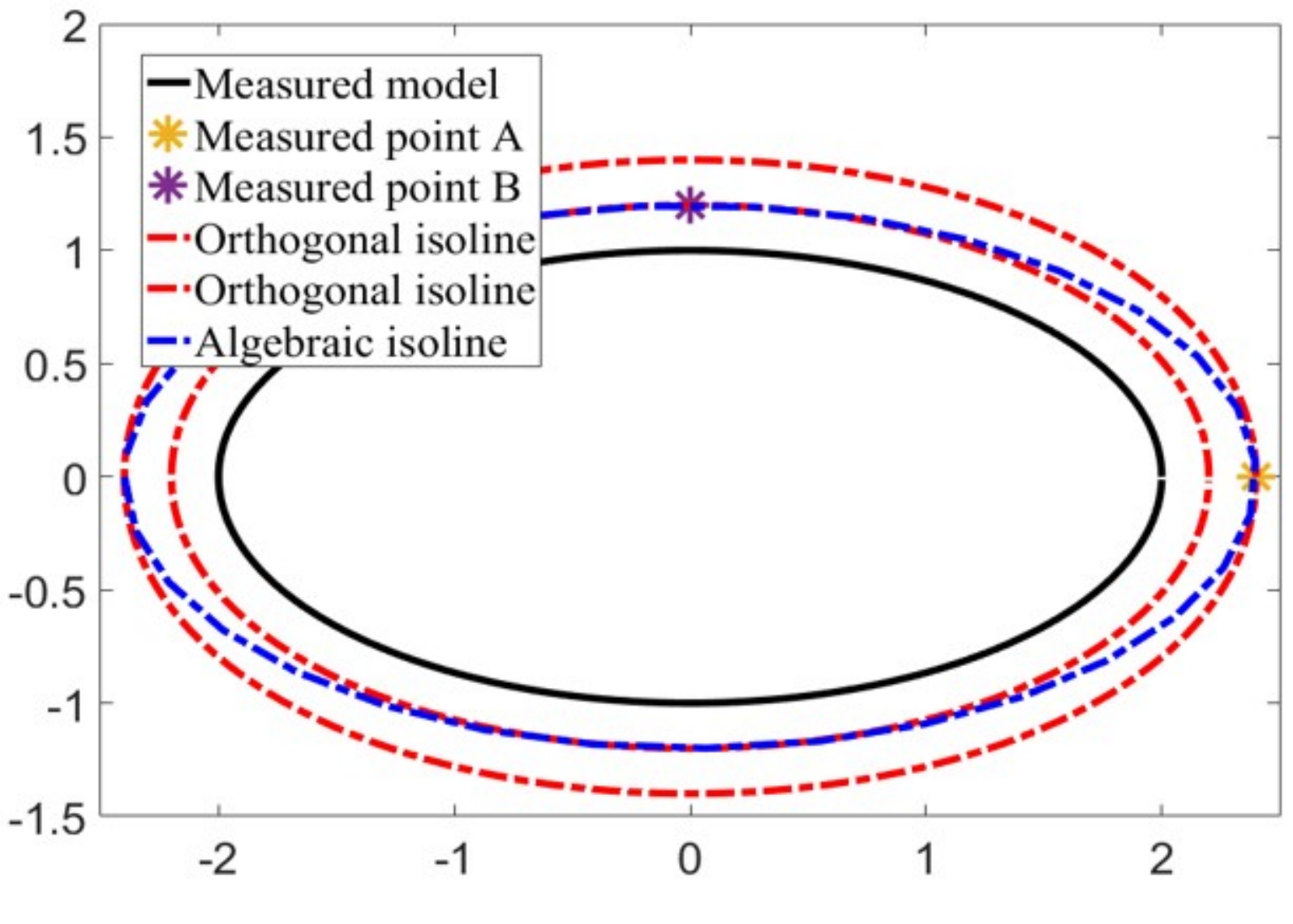}
\caption{Difference between the algebraic isoline and orthogonal isoline.}
\label{fig_1}
\end{figure}

The isoline of algebraic distance has the same shape as the measured model. Fig. 1 shows the difference between the isoline of the orthogonal distance and that of the algebraic distance. There exists complementarity between the algebraic distance and orthogonal distance from the view of metric. The complementarity of the two distances is proved by following an example. Fig. 1 shows that the algebraic distance of point A is equal to that of point B whereas the orthogonal distance of point A is larger than that of point B, which indicates that the algebraic distance may not increase (even decrease) when the orthogonal distance increases. Therefore, the combination of the two distances is a stricter metric and provides more constraints than a single distance to indicate that the quality (called point energy in sample consensus) of point A is lower than that of point B.

\section{Proposed Method}
\begin{figure*}[!t]
  \centering
  \includegraphics[width=1 \linewidth]{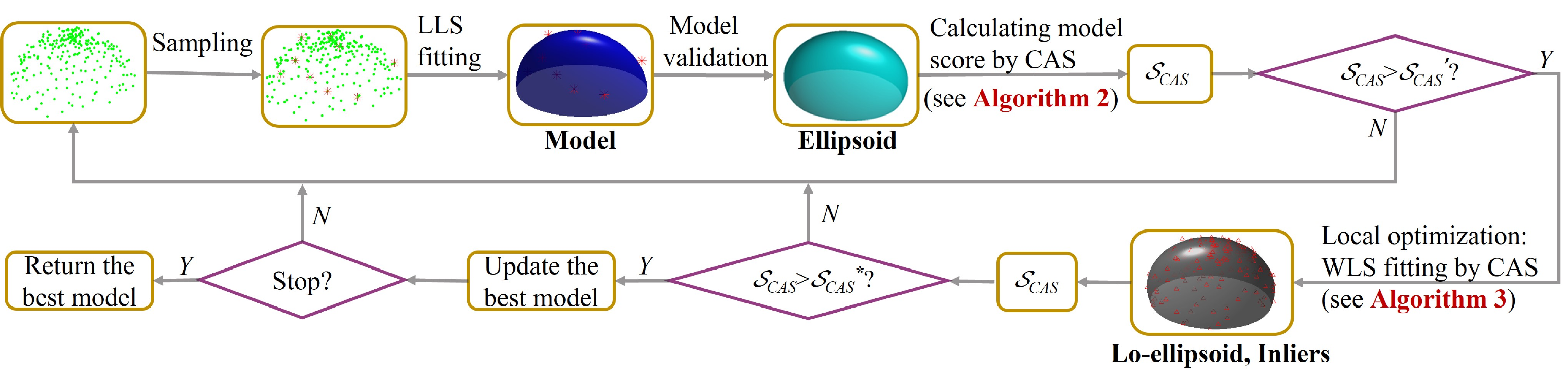}
  \caption{SC-based ellipsoid fitting using CAS. The Lo-ellipsoid is an ellipsoid from the local optimization step. ${\mathcal{S}_{CAS}}({\mathcal{S}_{CAS}}^\prime ,{\mathcal{S}_{CAS}}^{\text{*}})$ is the model score (the highest model score from RANSAC step, the highest model score). The strict metric (i.e., CAS) is used to calculate the model score and the weight of each point in WLS fitting. The proposed method shares a similar pipeline with sample-consensus-based methods.}
  \label{fig_2}
  \end{figure*}

Fig. 2 shows the architecture of the proposed ellipsoid fitting method. The proposed method is based on the sample consensus principle and contains a local optimization using WLS fitting. The major novelty is that both the model score and the weight in the WLS fitting are accurately calculated using CAS. The Sampson distance and axial distance are combined because 1) the algebraic distance has nongeometric limitations and is difficult to combine with other geometrical distances, 2) the calculation of the orthogonal distance is excessively time-consuming, 3) CAS is a strict metric and can provide strong constraint for the accurate model evaluation and WLS fitting.

\subsection{Scaling Factor}

\begin{figure}[!t]
  \centering
  \includegraphics[width=1 \linewidth]{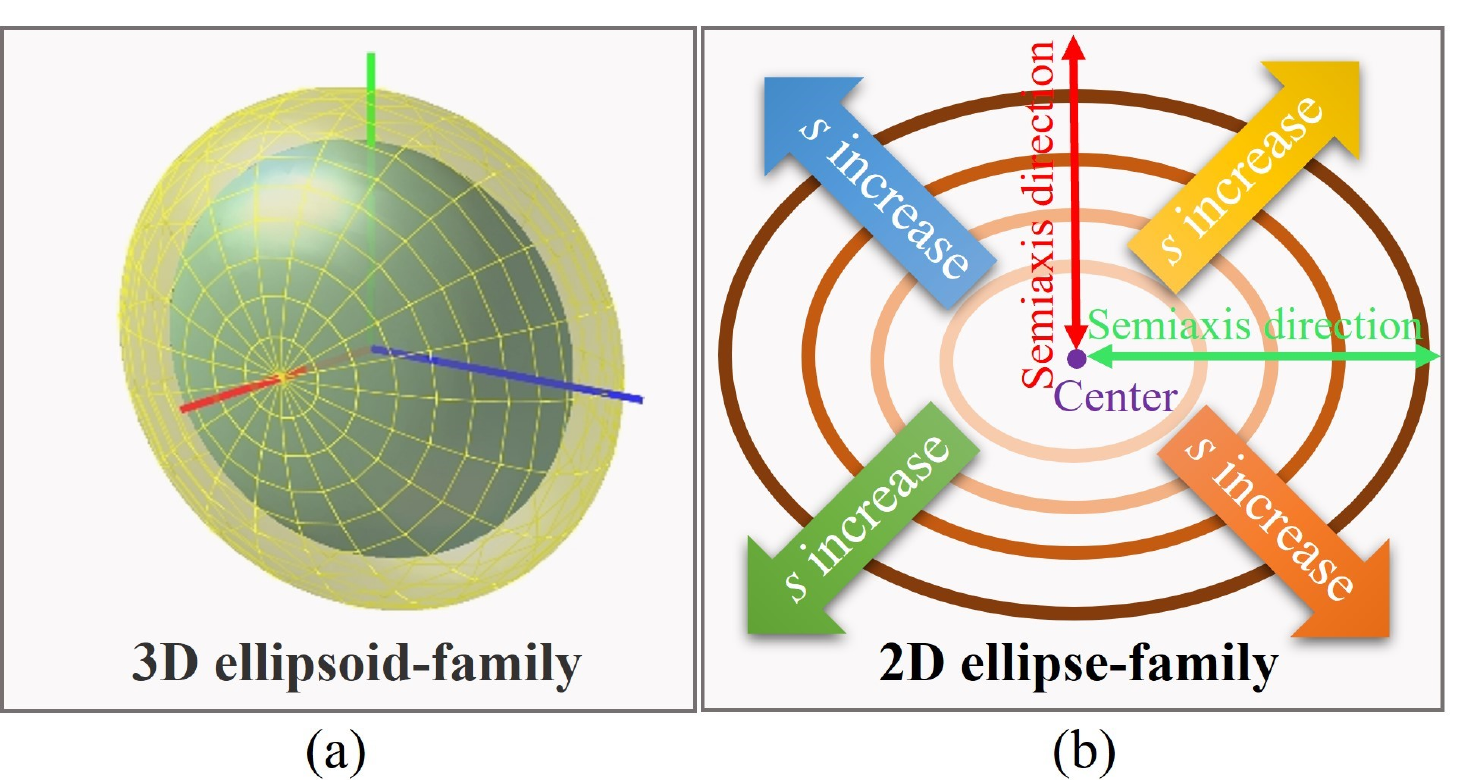}
  \caption{Two examples of the family. (a) 3-D ellipsoid family. (b) 2-D ellipse family.}
  \label{fig_3}
  \end{figure}

The axial distance is converted from the algebraic distance by using the scaling factor. Therefore, the notion of the scaling factor is introduced before the axial distance. Fig. 3(a) shows two members in the same ellipsoid family. For convenience, 2-D ellipses are used to introduce the family \cite{A8,A25}. Fig. 3(b) shows that a family consists of innumerable members that have the same shape (i.e., equal semiaxis ratio), same center and same semiaxis direction, but different semiaxis lengths. These members can be regarded as scaled ellipsoids obtained by scaling the semiaxis length of a member (called original ellipsoid ${M_1}$ which is generated by LLS fitting or WLS fitting).

Assuming that ${M_s}$ denotes a member with the scaling factor $s$, the ellipsoid family can be denoted by $\left\{ {{M_s}} \right\}$, $s \in [0, + \infty )$. Suppose that ${{\mathbf{r}}_s} = \left[ {r_s^{(1)},r_s^{(2)},r_s^{(3)}} \right]$ denotes a vector that includes three semiaxis lengths of the 3-D ellipsoid-member ${M_s}$. There will be
\begin{eqnarray}\label{eq5}
{{\mathbf{r}}_s} = s \cdot {{\mathbf{r}}_1},{\text{ }}s.t.{\text{ }}s \in [0, + \infty ).
 \end{eqnarray}

Particularly, a member with $s = 1$ is the original member ${M_1}$ of the family, $s > 1$ indicates that the semiaxis of a member is longer than that of the original member, $0 < s < 1$ indicates that the semiaxis of the member is shorter than that of the original member, and $s = 0$ indicates that the member degenerates into the center of the original member. Obviously, these members are the isosurfaces (isolines for 2-D ellipse) of algebraic distance when the original ellipsoid is a measured model, and the scaling factor $s$ reflects the magnitude of the algebraic distance.

\subsection{Axial Distance}
Although there exists complementarity between the algebraic distance and Sampson distance, the nongeometric property of algebraic distance results in some limitations \cite{A7} and prevents distance combination. To solve the nongeometric problems and realize the combination of complementary distances, the algebraic distance is converted to a geometrical axial distance through the scaling factor.

\begin{figure}[!t]
  \centering
  \includegraphics[width=1 \linewidth]{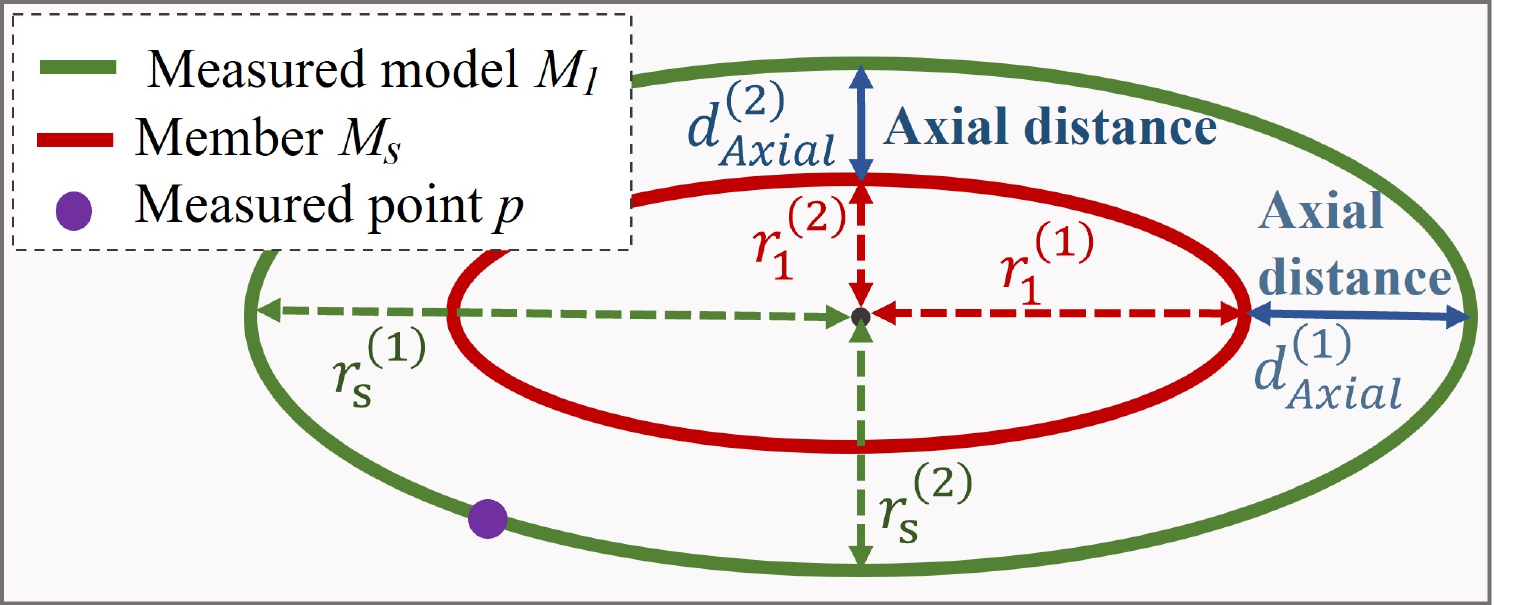}
  \caption{Axial distance between a measured point and measured model.}
  \label{fig_4}
  \end{figure}

Fig. 4 shows the measurement principle of axial distance between the measured model ${M_1}$ and measured point $p$ on a certain member ${M_s}$. An ellipse has two axial distances because it has two semiaxes. Similarly, a 3-D ellipsoid model has three axial distances. The $i$-th axial distance from the measured point $p$ to the measured model ${M_1}$ is defined as the difference of the $i$-th semiaxis of the measured model ${M_1}$ and that of the member ${M_s}$:
\begin{eqnarray}\label{eq6}
  d_{Axial}^{(i)}\left( {p,{M_1}} \right) = \left| {r_s^{(i)} - r_1^{(i)}} \right| = \left| {s - 1} \right| \cdot r_1^{(i)},
\end{eqnarray}

\noindent where $r_1^{(i)}$ and $r_s^{(i)}$ are the $i$-th semiaxis of the measured model ${M_1}$ and that of the member ${M_s}$. The axial distance satisfies the properties of positivity and symmetry, and it is a type of geometric distance. For convenience, three axis distances are merged into a value by L2 norm ${\left\|  \cdot  \right\|_2}$. Finally, the axial distance ${d_{Axial}}$ between the measured point $p$ and measured model ${M_1}$ is defined as
\begin{eqnarray}\label{eq7}
 \begin{gathered}
    {d_{Axial}}\left( {p,{M_1}} \right) = \frac{{\sqrt {d{{_{Axial}^{(1)}}^2} + d{{_{Axial}^{(2)}}^2} + d{{_{Axial}^{(3)}}^2}} }}{3} \\ 
     = \left| {s - 1} \right| \cdot \frac{{{{\left\| {{{\mathbf{r}}_1}} \right\|}_2}}}{3}. \\ 
  \end{gathered}
\end{eqnarray}

Additionally, Fig. 4 also shows that the member ${M_s}$ is the isoline of axial distance. Meanwhile, the member ${M_s}$ is also the isoline of algebraic distance. The isolines of the algebraic distance and axial distance have the same shape as the measured model ${M_1}$. Therefore, the measurement ways of the axial distance and algebraic distance are similar. However, the geometric axial distance is superior to the algebraic distance, which has nongeometric problems.
      
For a member ${M_s}$, ${{\mathbf{Q}}_s} \in {\mathbb{R}^{4 \times 4}}$ denotes its parameter matrix, ${{\mathbf{q}}_s} \in {\mathbb{R}^{10 \times 1}}$ is the vector form of ${{\mathbf{Q}}_s}$, and the subscript $s$ (if any) denotes its scaling factor. The ellipsoid family fills out the entire 3-D space. Thus, a measured point $p$ must be located on a certain member ${M_s}$. The homogenous coordinate ${{\mathbf{x}}_h}$ of the point $p$ on the member ${M_s}$ satisfies
\begin{eqnarray}\label{eq8}
{\mathbf{x}}_h^T{{\mathbf{Q}}_s}{{\mathbf{x}}_h} = 0,{\text{ }}s.t.{\text{ }}s \in [0, + \infty ).
\end{eqnarray}

Before calculating the axial distance between the point $p$ and model ${M_1}$, the scaling factor $s$ of the member ${M_s}$ through the measured point $p$ must be solved by (8). However, the parameter matrix ${{\mathbf{Q}}_s}$ in (8) is unknown. The measured ellipsoid ${M_1}$ is generated by LLS fitting or WLS fitting; therefore, the parameter matrix ${{\mathbf{Q}}_1}$ is known. To solve the scaling factor $s$, the parameter matrix ${{\mathbf{Q}}_s}$ of the entire ellipsoid family is derived by using the known ${{\mathbf{Q}}_1}$ as follows.

Assume that ${{\mathbf{\vec Q}}_s} \in {\mathbb{R}^{4 \times 4}}$ denotes the standard form of parameter matrix ${{\mathbf{Q}}_s}$. The standard form ${{\mathbf{\vec Q}}_s}$ is expressed as
\begin{eqnarray}\label{eq9}
{{\mathbf{\vec Q}}_s} = \left[ {\begin{array}{*{20}{c}}
  {\psi({{\mathbf{r}}_s})}&{\mathbf{0}} \\ 
  {{{\mathbf{0}}^T}}&{ - 1} 
\end{array}} \right],
\end{eqnarray}

\noindent where the operator $\psi\left(  \cdot  \right)$ converts a vector to a diagonal matrix whose diagonal elements are the inverse of squares of the vector elements. The relationship between the standard form ${{\mathbf{\vec Q}}_s}$ and the parameter matrix ${{\mathbf{Q}}_s}$ is 
\begin{eqnarray}\label{eq10}
  {{\mathbf{Q}}_s} = {{\mathbf{H}}^T}{{\mathbf{\vec Q}}_s}{\mathbf{H}} = \frac{1}{{{s^2}}}\left[ {\begin{array}{*{20}{c}}
    {{{\mathbf{R}}^T}\psi ({{\mathbf{r}}_1}){\mathbf{R}}}&{{{\mathbf{R}}^T}\psi ({{\mathbf{r}}_1}){\mathbf{T}}} \\ 
    {{{\mathbf{T}}^T}\psi ({{\mathbf{r}}_1}){\mathbf{R}}}&{{{\mathbf{T}}^T}\psi ({{\mathbf{r}}_1}){\mathbf{T}}{\text{ - }}{s^2}} 
  \end{array}} \right]
\end{eqnarray}

\noindent where ${\mathbf{H}} = [{\mathbf{R}},{\mathbf{T}};{\text{ }}{{\mathbf{0}}^T}{\text{,}}1]$ denotes a Euclidean transformation matrix, ${\mathbf{R}} \in {\mathbb{R}^{3 \times 3}}$ is a rotation matrix, and ${\mathbf{T}} \in {\mathbb{R}^{3 \times 1}}$ is a translation matrix. All members share the same Euclidean transformation matrix ${\mathbf{H}}$ because they share the same center and same semiaxis direction.
The original ellipsoid ${M_1}$ has been generated by using LLS or WLS. Therefore, the parameter matrix ${{\mathbf{Q}}_1}$ is known and can be expressed as
\begin{eqnarray}\label{eq11}
  {{\mathbf{Q}}_1} = \left[ {\begin{array}{*{20}{c}}
    {{\mathbf{U\Lambda }}{{\mathbf{U}}^T}}&{{\mathbf{q}}_{_1}^{\left( {7 - 9} \right)}} \\ 
    {{{\left( {{\mathbf{q}}_{_1}^{\left( {7 - 9} \right)}} \right)}^T}}&{{\mathbf{q}}_{_1}^{\left( {10} \right)}} 
  \end{array}} \right],
\end{eqnarray}

\noindent where the diagonal matrix ${\mathbf{\Lambda }}$ and orthogonal matrix ${\mathbf{U}}$ are obtained by using the eigenvalue decomposition (EVD) of the upper-left $3 \times 3$ submatrix ${\mathbf{Q}}_1^{3 \times 3}$  of ${{\mathbf{Q}}_1}$, ${\mathbf{q}}_{_1}^{\left( {7 - 9} \right)}$ is a vector that is composed of 7th–9th elements from ${{\mathbf{q}}_1}$, and ${\mathbf{q}}_{_1}^{\left( {10} \right)}$ is the 10th element of ${{\mathbf{q}}_1}$. The original ellipsoid ${M_1}$ is also a member of the family $\left\{ {{M_s}} \right\}$. Substituting $s = 1$ into (10) gives
\begin{eqnarray}\label{eq12}
{{\mathbf{Q}}_1} = \left[ {\begin{array}{*{20}{c}}
  {{{\mathbf{R}}^T}\psi({{\mathbf{r}}_1}){\mathbf{R}}}&{{{\mathbf{R}}^T}\psi({{\mathbf{r}}_1}){\mathbf{T}}} \\ 
  {{{\mathbf{T}}^T}\psi({{\mathbf{r}}_1}){\mathbf{R}}}&{{{\mathbf{T}}^T}\psi({{\mathbf{r}}_1}){\mathbf{T}}{\text{ - }}1} 
\end{array}} \right].
\end{eqnarray}

It should be noted that both ${\mathbf{Q_1}}$ and $l \cdot {\mathbf{Q_1}}$ represent the same model for an arbitrary scalar $l$. Comparing (11) to (12) gives
\begin{eqnarray}\label{eq13}
  \left[ \begin{gathered}
    {\mathbf{U\Lambda }}{{\mathbf{U}}^T}\ \ \ \ \ {\mathbf{q}}_{_1}^{\left( {7 - 9} \right)} \hfill \\
    {\left( {{\mathbf{q}}_{_1}^{\left( {7 - 9} \right)}} \right)^T}{\text{ }}{\mathbf{q}}_{_1}^{\left( {10} \right)} \hfill \\ 
  \end{gathered}  \right] \! = \! l\left[ \begin{gathered}
    {{\mathbf{R}}^T}\psi ({{\mathbf{r}}_1}){\mathbf{R}}\ \ \ {{\mathbf{R}}^T}\psi ({{\mathbf{r}}_1}){\mathbf{T}} \hfill \\
    {{\mathbf{T}}^T}\psi ({{\mathbf{r}}_1}){\mathbf{R}}\ \ \ {{\mathbf{T}}^T}\psi ({{\mathbf{r}}_1}){\mathbf{T}}{\text{ - }}1 \hfill \\ 
  \end{gathered}  \right].
\end{eqnarray}

We can add a constraint of ${\mathbf{U}} = {{\mathbf{R}}^T}$ because both ${\mathbf{U}}$ and ${\mathbf{R}}$ are the orthogonal matrix. According to (13), the elements of the parameter matrix ${{\mathbf{Q}}_s}$ can be solved by
\begin{eqnarray}\label{eq14}
  \left\{ \begin{gathered}
    {\text{ }}{\mathbf{R}} = {{\mathbf{U}}^T} \hfill \\
    {\text{ }}{\mathbf{T}} = {{\mathbf{\Lambda }}^{ - 1}}{{\mathbf{U}}^T}{\mathbf{q}}_{_1}^{\left( {7 - 9} \right)} \hfill \\
    {\text{ }}\psi\left( {{{\mathbf{r}}_1}} \right) = {{\mathbf{\Lambda }} \mathord{\left/
   {\vphantom {{\mathbf{\Lambda }} {\left( {{\mathbf{q}}{{_{_1}^{\left( {7 - 9} \right)}}^T}{\mathbf{U}}{{\mathbf{\Lambda }}^{{\text{ - }}1}}{{\mathbf{U}}^T}{\mathbf{q}}_{_1}^{\left( {7 - 9} \right)} - {\mathbf{q}}_{_1}^{\left( {10} \right)}} \right)}}} \right.
   \kern-\nulldelimiterspace} {\left( {{\mathbf{q}}{{_{_1}^{\left( {7 - 9} \right)}}^T}{\mathbf{U}}{{\mathbf{\Lambda }}^{{\text{ - }}1}}{{\mathbf{U}}^T}{\mathbf{q}}_{_1}^{\left( {7 - 9} \right)} - {\mathbf{q}}_{_1}^{\left( {10} \right)}} \right)}} \hfill \\ 
  \end{gathered}  \right..
\end{eqnarray}

Thus far, the ellipsoid family with a physical interpretability was built by (10) and (14) based on the known original ellipsoid ${M_1}$ and the scaling factor. We substitute (10) into (8) to solve the scaling factor s of the member ${{\mathbf{Q}}_s}$ through the measured point $p$:
\begin{eqnarray}\label{eq15}
  \begin{gathered}
    {s^2} = {\mathbf{x}}_h^T\left[ {\begin{array}{*{20}{c}}
    {{{\mathbf{R}}^T}\psi({{\mathbf{r}}_1}){\mathbf{R}}}&{{{\mathbf{R}}^T}\psi({{\mathbf{r}}_1}){\mathbf{T}}} \\ 
    {{{\mathbf{T}}^T}\psi({{\mathbf{r}}_1}){\mathbf{R}}}&{{{\mathbf{T}}^T}\psi({{\mathbf{r}}_1}){\mathbf{T}}} 
  \end{array}} \right]{{\mathbf{x}}_h},{\text{ }} \hfill \\
    {\text{                                         }}s.t.{\text{ }}s \in [0, + \infty ). \hfill \\ 
  \end{gathered}
\end{eqnarray}

Finally, the axial distance between the measured point $p$ and the measured ellipsoid ${M_1}$ can be calculated by using the solved scaling factor $s$ and (7).

\subsection{Combination of Complementary Distances}
The axial distance is similar to the algebraic distance, which is complementary to the orthogonal distance and Sampson distance. Therefore, the axial distance is also complementary to the orthogonal distance and Sampson distance. In this section, the complementarity between the axial distance and orthogonal distance is further proved from the perspective of fitting error. Subsequently, the reason for selecting the axial distance and Sampson distance in the combination is explained. Finally, the distances are combined in a manner of weighted sum.

\begin{figure}[!t]
  \centering
  \includegraphics[width=1 \linewidth]{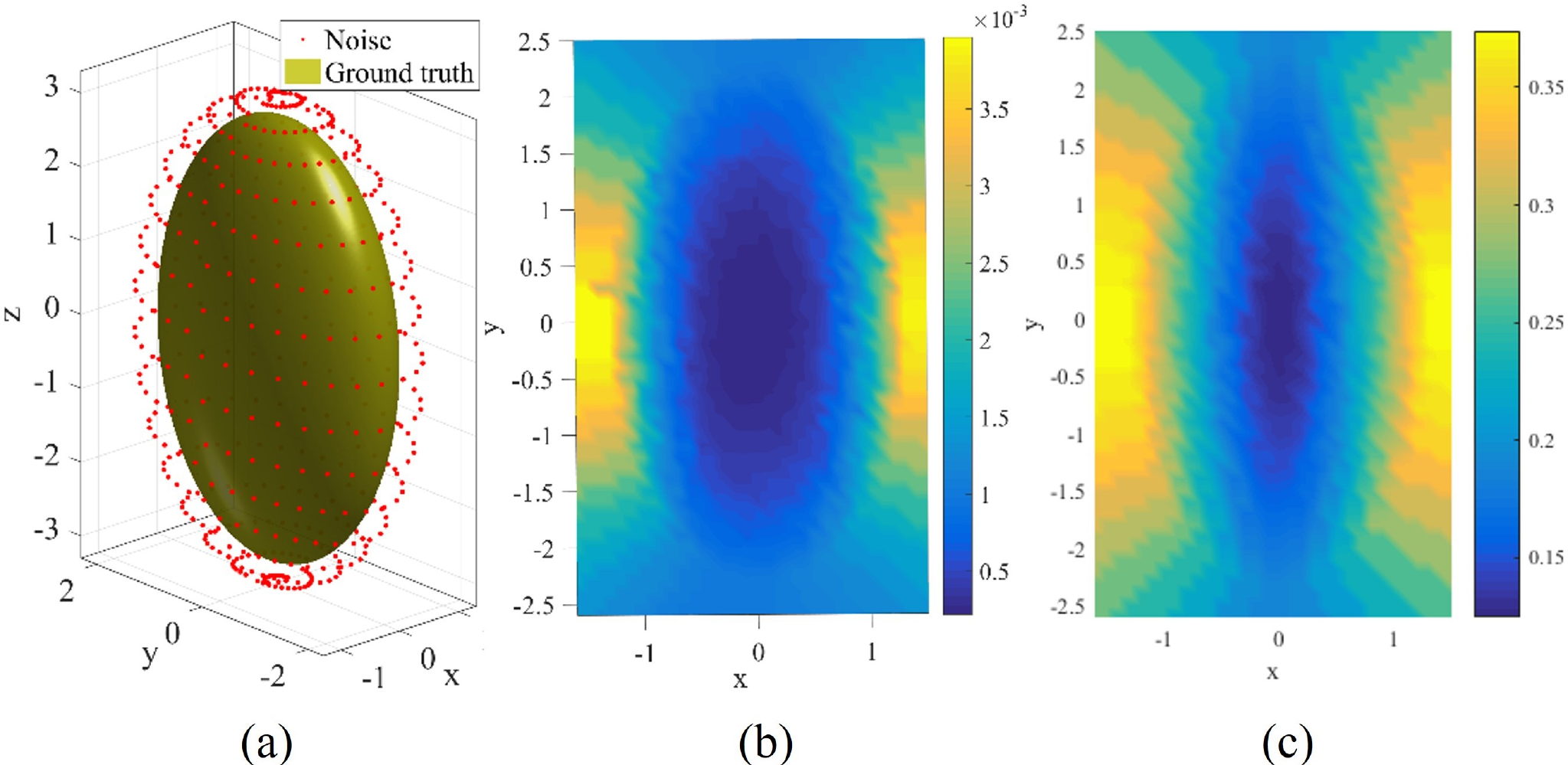}
  \caption{Effect of noise on LLS fitting. (a) Noise points on an orthogonal isosurface. (b) Fitting error. (c) Axial distances between the noise points and the true ellipsoid.}
  \label{fig_5}
  \end{figure}

In the sample consensus, the noise level of the measured point directly affects the point energy contributed to model score and its weight in WLS fitting. Therefore, the evaluation of noise level is crucial for the model evaluation and WLS fitting. A good point-to-model distance should be able to accurately evaluate the noise level of a measured point. Generally, the fitting error caused by a point denotes the noise level. The complementarity of the distances is further proved using the fitting error in the following test. Fig. 5(a) shows a true ellipsoid and noise points on an orthogonal isosurface (called the isoline in 2-D space). In the test, one noise point was taken at one time to interfere with LLS fitting with the points on the true ellipsoid.

Fig. 5(b) shows the fitting errors which were caused by noise points and were plotted at X-Y coordinates of each noise point. The results show that the fitting error varies with the position of the noise point relative to the true ellipsoid. The closer to the steep section of the true ellipsoid the noise point is, the smaller the error will be. In contrast, the closer to the flat section of the true ellipsoid the noise point is, the larger the error will be. In summary, the noise level of noise points on orthogonal isosurface is related to the shape of the true ellipsoid. Therefore, the distance metric whose isosurfaces have the same shape as the true ellipsoid can accurately evaluate the noise level of points on the orthogonal isosurface.
 
Fig. 5(c) shows that the amplitude distribution of the axial distances of noise points on the orthogonal isosurface is similar to that of the fitting error. The result indicates that the axial distance can accurately evaluate the noise level of noise points on the orthogonal isosurface. However, the orthogonal distance fails to distinguish the noise points on the orthogonal isosurface. Therefore, there is a complementarity between the axial distance and orthogonal distance because the distance combination can form a novel stricter metric.

To form the strict metric, a linear distance combination is proposed. The Sampson distance is the first-order approximation of the orthogonal distance. The axial distance and algebraic distance are similar because the axial distance is converted from the algebraic distance and their isolines are same-shape as the measured model. According to the metric difference of distances, the listed distances can be divided into two categories. The first category includes the Sampson distance and orthogonal distance, and the second category includes the algebraic distance and axial distance. The distances in different categories are complementary whereas the distances in the same category are non-complementary. The computation complexity of Sampson distance are significantly smaller than that of the orthogonal distance. The algebraic distance has nongeometric limitations and is difficult to combine with the geometric Sampson distance. Based on the above analysis, the axial distance and Sampson distance are combined.

Assume that the algebraic equation of a measured model $M$ is denoted by $F\left( {\mathbf{x}} \right)$. Based on the definition [25], the Sampson distance ${d_s}$ is solved by
\begin{eqnarray}\label{eq16}
{\text{ }}{d_{Sampson}}\left( {p,M} \right) = \left| {\frac{{F\left( {{{\mathbf{x}}_h}} \right)}}{{{{\left\| {\nabla F\left( {{{\mathbf{x}}_h}} \right)} \right\|}_2}}}} \right|,
\end{eqnarray}
\noindent where $\nabla$ denotes a gradient operator. We implement the distance combination in the manner of weighted sum. The combination between the axial distance and Sampson distance (CAS) ${d_{CAS}}$  are expressed as
\begin{eqnarray}\label{eq17}
{d_{CAS}}\left( {p,M} \right) \!=\! \lambda{d_{Axial}}\left( {p,M} \right) \!+\! \left( {1 \!\! - \!\! \lambda } \right){d_{Sampson}}\left( {p,M} \right),
\end{eqnarray}
\noindent where $\lambda$ is the control ratio of distance combination. Two types of distances are combined into a strict metric when $0 < \lambda  < 1$. It is recommended to set $\lambda$ to 0.5 to put the two distances on the same order of magnitude.

\subsection{Ellipsoid Fitting by CAS}

\begin{table}[!t]
  \centering
  \includegraphics[width=1 \linewidth]{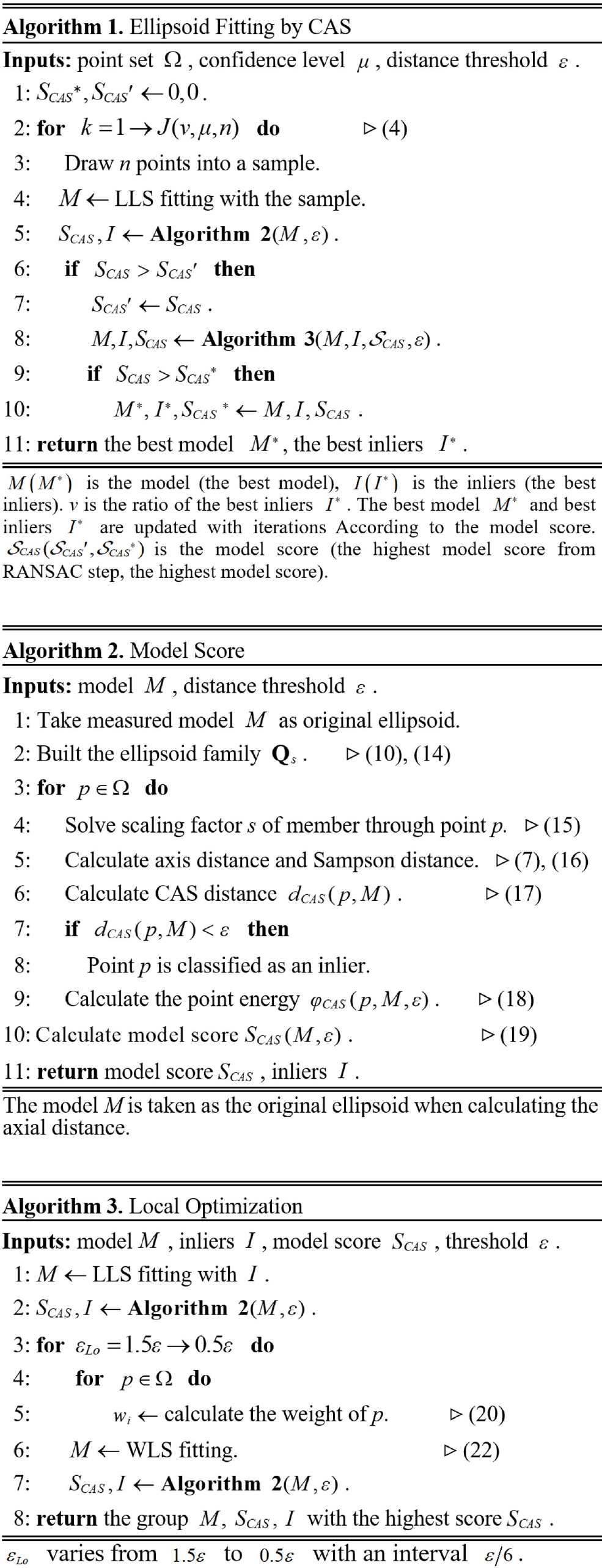}
  \vspace{-20 pt}
\end{table}

In this section, a novel ellipsoid fitting method is introduced. The proposed method is based on the sample consensus. CAS is employed as the point-to-model distance to utilize the strict metric. \textbf{Algorithm 1} shows the pipeline of the proposed method, \textbf{Algorithm 2} introduces the calculation method of model score, and \textbf{Algorithm 3} introduces a local optimization using WLS fitting.

\textbf{Algorithm 1} contains a RANSAC step. Sample consensus searches for the best model among the models
generated by using model fitting. The minimal subset sampling strategy, which draws nine points ($n = 9$) into a sample to determine an ellipsoid, is adopted. In each iteration, a model is generated by LLS fitting with a sample. The required number of iterations $J(v,\mu ,n)$ is updated according to (4). After the required iterations are performed, some models are generated from the RANSAC and local optimization steps. The best model ${M^{\text{*}}}$ is found by maximizing the model score ${\mathcal{S}_{CAS}}$.

\textbf{Algorithm 2} shows the calculation method of the model score ${\mathcal{S}_{CAS}}\!\left( {M,\varepsilon } \right)$ by using CAS. First, the axial distance ${d_{Axial}}\left( {p,M} \right)$ is calculated by taking the measured model $M$ as the original ellipsoid ${M_1}$. Subsequently, the CAS distance ${d_{CAS}}\left( {p,M} \right)$ from a point $p$ to measured model $M$ is calculated by using (17). The point energy ${\varphi _{CAS}}$ of the point $p$ is
\begin{eqnarray}\label{eq18}
{\varphi _{CAS}} \!\left( {p,M,\varepsilon } \right) = \exp \left( { - \frac{{{d_{CAS}}{{\left( {p,M} \right)}^2}}}{{2{\varepsilon ^2}}}} \right).
\end{eqnarray}
Finally, the model score ${\mathcal{S}_{CAS}}\!\left( {M,\varepsilon } \right)$ of the model $M$ is
\begin{eqnarray}\label{eq19} 
{\mathcal{S}_{CAS}}\!\left( {M,\varepsilon } \right) = \sum\limits_{p \in \Omega } {{\varphi _{CAS}}} \!\left( {p,M,\varepsilon } \right).
\end{eqnarray}

\textbf{Algorithm 3} introduces the local optimization step, which is performed only when the best model from the RANSAC step is updated. The local optimization step uses WLS fitting method to optimize the best model from the RANSAC step. In conventional WLS fitting method, the weight of each point is calculated by using algebraic distance \cite{A19}. However, the proposed method uses CAS distance to calculate the weight in WLS fitting. Relative to the model $M$, the weight ${w_i}$ of $i$-th point $p$ is calculated as
\begin{eqnarray}\label{eq20}
{w_i} = \exp \left( { - \frac{{{d_{CAS}}{{\left( {p,M} \right)}^2}}}{{2{\varepsilon _{Lo}}^2}}} \right),
\end{eqnarray}
\noindent where the distance threshold ${\varepsilon _{Lo}}$ varies from 1.5$\varepsilon$ to 0.5$\varepsilon$ to reduce the sensitivity of fitting to the distance threshold ${\varepsilon}$. Assume that $C$ denotes the total number of points in the input $\Omega$, and ${[{x_1},{x_2},{x_3}]^T}$ denotes the coordinate of the $i$-th point. WLS fitting estimates the model parameters ${\mathbf{q}} \in {\mathbb{R}^{10 \times 1}}$ by minimizing a residual vector ${{\mathbf{r}}_{CAS} = [{r_1},{r_2},...,{r_C}]^T}$ where the $i$-th residual ${r_i}$ is expressed as
\begin{eqnarray}\label{eq21}
{r_i} = {w_i} \cdot {{{\mathbf{d}}_i}^T}{\mathbf{q}},
\end{eqnarray}
\noindent and\\
 ${{\mathbf{d}}_i} \!= \! {[{x_1}^2,{x_2}^2,{x_3}^2,2{x_1}{x_2},2{x_1}{x_3},2{x_2}{x_3},2{x_1},2{x_2},2{x_3}, -1]^T}$. To estimate the model parameters ${\mathbf{q}}$, WLS fitting method solves
\begin{eqnarray}\label{eq22}
  \mathop {\min }\limits_{\mathbf{q}} \left\| {{{\mathbf{r}}_{CAS}}} \right\|_2^2 =  \mathop {\min }\limits_{\mathbf{q}} \left( {{{\mathbf{q}}^T}{\mathbf{D}}{{\mathbf{D}}^T}{\mathbf{q}}} \right),{\text{ }}s.t.{\text{ }}\left\| {\mathbf{q}} \right\| = 1,
\end{eqnarray}
\noindent where ${\mathbf{D}} = [{w_1} \cdot {{\mathbf{d}}_1},...,{w_C} \cdot {{\mathbf{d}}_C}] \in {\mathbb{R}^{10 \times C}}$. The solution of WLS fitting is an eigenvector corresponding to the smallest eigenvalue of ${\mathbf{D}}{{\mathbf{D}}^T}$.

Additionally, the ellipsoid solution may not be guaranteed by LLS fitting and WLS fitting. The model validation is performed immediately after a model is generated to search the ellipsoid solution by judging the positive semi-definiteness of matrix \cite{A9}. If model validation fails, the algorithm will directly jump to the next iteration. The proposed method performs the point classification by using CAS. Because CAS distance is geometric, the distance threshold $\varepsilon$ can be set as a geometric value. The class label $L(p)$ of a point $p$ is
\begin{eqnarray}\label{eq23}
L\left( p \right) = \left\{ {\begin{array}{*{20}{c}}
  {inlier{\ \ \ \text{    if }\ }{d_{CAS}}\left( {p,M} \right) < \varepsilon } \\ 
  {outlier{\ \ \text{    if }\ }{d_{CAS}}\left( {p,M} \right) \geqslant \varepsilon } 
\end{array}} \right..
\end{eqnarray}

\section{Experiments}

The experiments were conducted on synthetic and real datasets. The proposed method was compared to direct ellipsoid fitting methods and robust SC-based fitting methods. The direct fitting methods include SOD \cite{A10} and ADMM \cite{A9}. The robust SC-based methods include RANSAC \cite{A15}, FLO \cite{A19}, and GC \cite{A20}. Additionally, we compare CAS distance with algebraic distance \cite{A8}, Sampson distance \cite{A25}, orthogonal distance \cite{A29}, and several combined distances.

The proposed method uses CAS to calculate the model score and weight in WLS fitting. By default, RANSAC, FLO, and GC are implemented by using the Sampson distance because of its popularity. For a fair comparison, all SC-based methods adopt the Gaussian kernel to normalize the point energy to 0–1. GC is implemented using C++ and other methods are implemented in MATLAB 2022. The experiments ran on a computer with a 2.5 GHz processor and 8 GB RAM. In the initialization stage, the control ratio $\lambda$ in CAS was set to 0.5 to put the two distances on the same order of magnitude. For fair comparison, the confidence levels $\mu$ of all SC-based methods were set to 0.95. The distance threshold $\varepsilon$ was tuned to achieve the highest accuracy. The direct fitting methods were initialized as reported in \cite{A9,A10}.

\begin{figure}[!t]
  \centering
  \includegraphics[width=1 \linewidth]{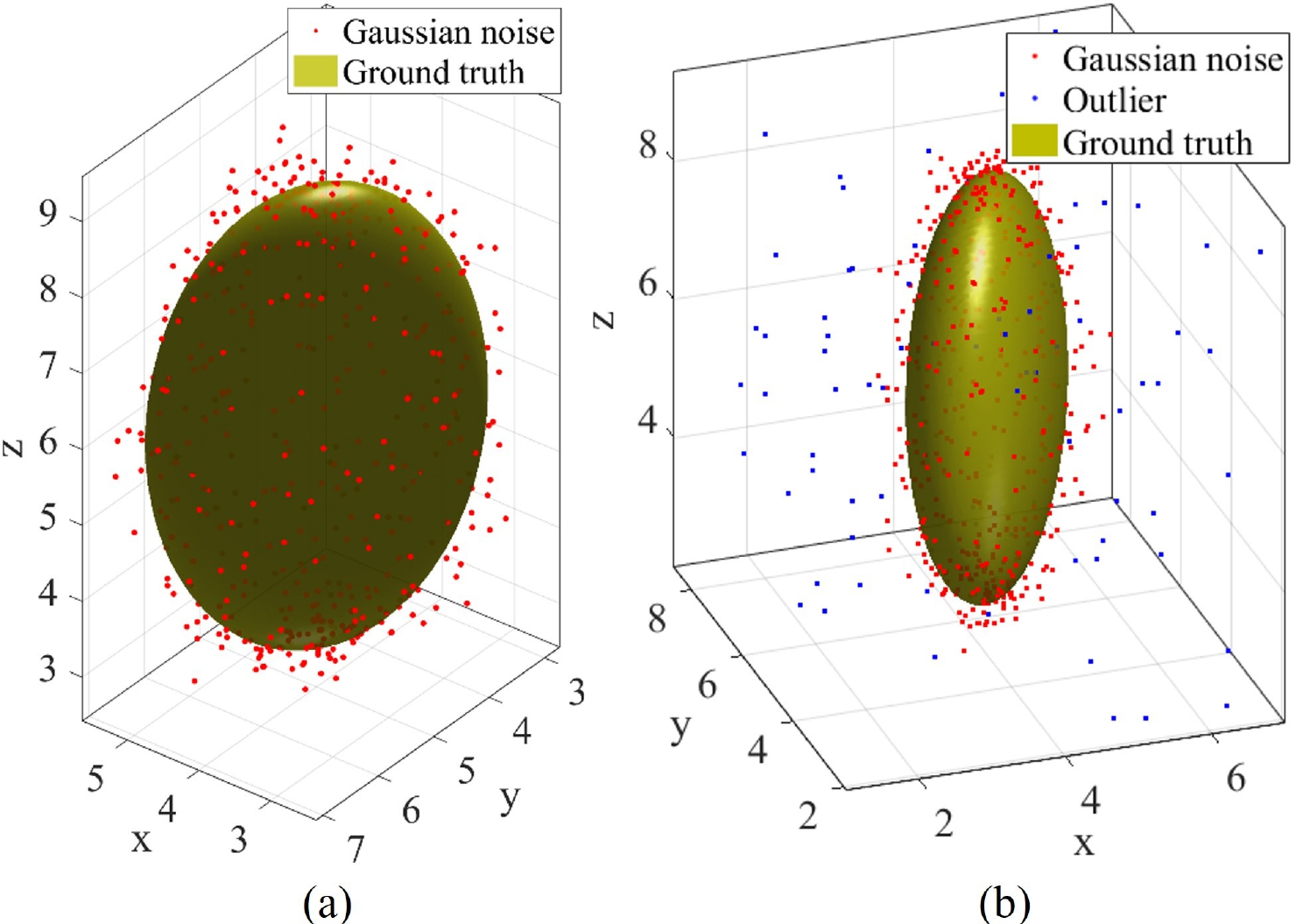}
  \caption{Instances of (a) Gaussian-noise dataset and (b) outlier dataset.}
  \label{fig_6}
  \end{figure}

The synthetic dataset consists of a Gaussian-noise dataset and an outlier dataset. Fig. 6 shows instances of the Gaussian-noise and outlier datasets. True ellipsoids were randomly generated. For the Gaussian-noise dataset, the zero-mean Gaussian noise was added to the coordinates of 500 points on the true ellipsoid, and the standard deviation $\sigma$ varied from 0.1 to 0.4 times of the semiaxis of ellipsoids. For the outlier dataset, the standard deviation $\sigma$ of zero-mean Gaussian noise was set to 0.25, and a certain proportion of outliers were added. The proportion of outliers varied from 10$\%$ to 40$\%$.

A total of 10 ellipsoid-instances were created at each noise level. 100 runs were performed on each instance. Therefore, 1000 runs were performed at each noise level. The mean values of three types of errors were reported, respectively. The three types of errors are
\begin{eqnarray}\label{eq24}
\left\{ \begin{gathered}
  {\text{ Parameter Error}} = {\left\| {{{\mathbf{q}}_{estimated}}{\text{ - }}{{\mathbf{q}}_{true}}} \right\|_1} \hfill \\
  {\text{ Semiaxis Error}} = {\left\| {{{\mathbf{r}}_{estimated}}{\text{ - }}{{\mathbf{r}}_{true}}} \right\|_1} \hfill \\
  {\text{ Center Error}} = {\left\| {{{\mathbf{c}}_{estimated}}{\text{ - }}{{\mathbf{c}}_{true}}} \right\|_1} \hfill \\ 
\end{gathered}  \right.,
\end{eqnarray}

\noindent where ${\mathbf{q}}$ is the parameter vector of fitting result, ${\mathbf{r}}$ is the semiaxis vector, and ${\mathbf{c}}$ is the center vector. The subscripts denote the estimated value and the ground truth value. ${\left\|  \cdot  \right\|_1}$ is an operator of L1 norm.

\subsection{Comparison of Ellipsoid Fitting Methods}

\begin{figure*}[!t]
  \centering
  \includegraphics[width=1 \linewidth]{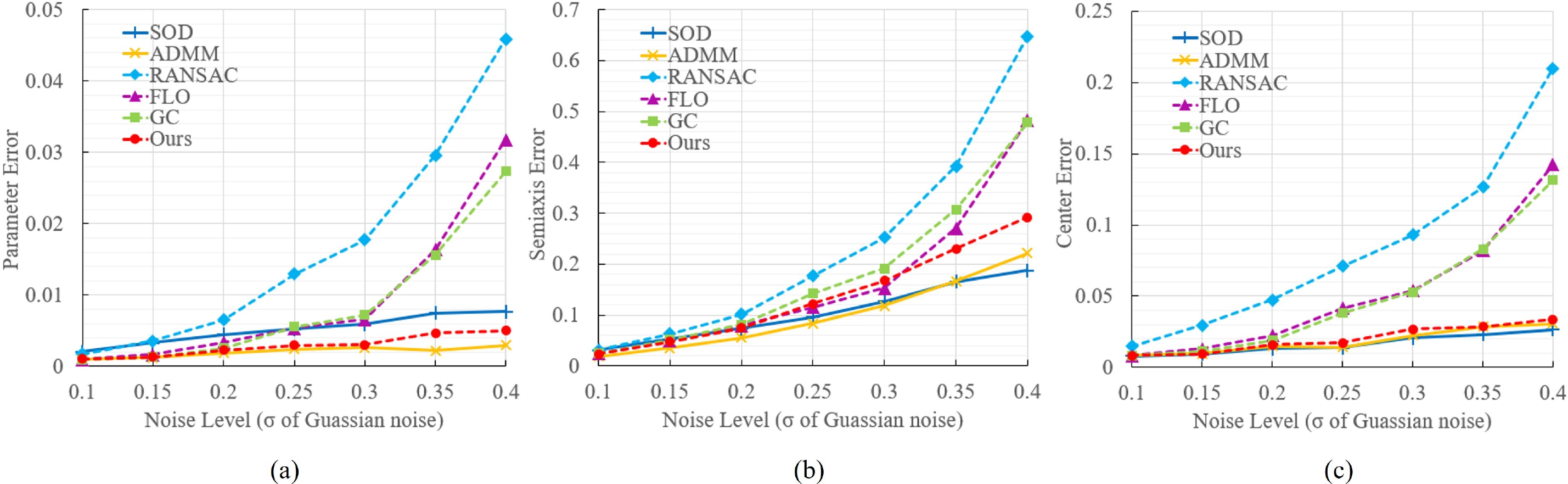}
  \caption{Fitting errors of the ellipsoid fitting methods on the Gaussian-noise dataset. (a) Parameter error. (b) Semiaxis error. (c) Center error. The noise level increases with the standard deviation $\sigma$. The dashed lines represent the SC-based methods whereas the solid lines represent the direct fitting methods.}
  \label{fig_7}
  \end{figure*}

1) \textbf{Accuracy on Gaussian-Noise Dataset:} As shown in Fig. 7, the accuracies of ellipsoid fitting methods are compared on the Gaussian-noise dataset. The results show that SOD and ADMM outperform RANSAC, FLO, and GC in the absence of outliers. Owing to the random sampling, sample consensus produces some low-accuracy fitting results in certain runs. The direct fitting methods achieve high accuracy. Although the proposed method is based on the sample consensus principle, it achieves high accuracy, which is close to that of the direct fitting method and is the highest among SC-based methods.

\begin{figure*}[!t]
  \centering
  \includegraphics[width=1 \linewidth]{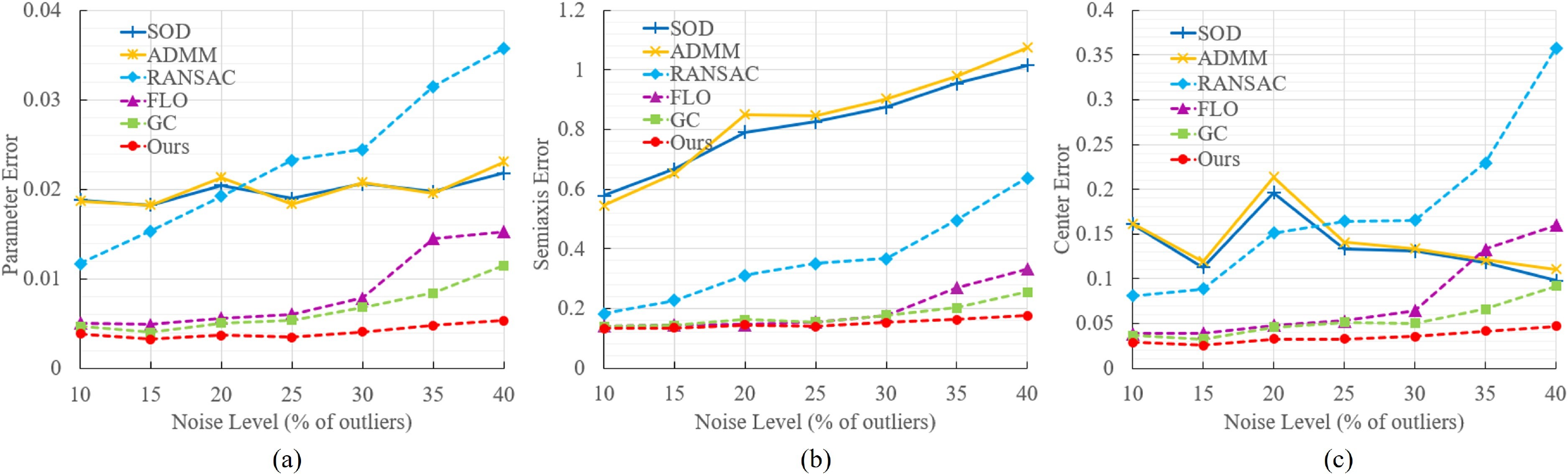}
  \caption{Fitting errors of the ellipsoid fitting methods on outlier dataset. (a) Parameter error. (b) Semiaxis error. (c) Center error. The noise level increases with the proportion ($\%$) of outliers.}
  \label{fig_8}
  \end{figure*}

2) \textbf{Accuracy on Outlier Dataset:} Fig. 8 shows that the proposed method achieves the highest accuracy in the presence of outliers. The SC-based methods are more robust against outliers than the direct fitting methods, which proves the advantage of robustness of the sample consensus principle. Owing to the high robustness against outliers, the proposed method overperformed the direct fitting methods by a large margin in terms of accuracy. We can compare the proposed method with other SC-based methods to show the accuracy improved by CAS rather than the one improved by the sample consensus. With the proportion of outliers increased, the proposed method achieves stable accuracy, which is the highest among SC-based methods. The results show that the proposed method further improves the robustness against outliers and achieves consistently high accuracy on both Gaussian-noise and outlier datasets.

\subsection{Comparison of Distance Metrics}
To study the effect of the distance metric on ellipsoid fitting, the proposed method was compared to the RANSAC methods using different types of distances. RANSAC was implemented with orthogonal, Sampson, algebraic, axial, Sampson + orthogonal, axial + orthogonal, and CAS distances (called CAS@RANSAC to differ from the proposed method). The combinations between algebraic distance and other distances are not shown because the algebraic distance is nongeometric and dimensionless.

\begin{figure*}[!t]
  \centering
  \includegraphics[width=1 \linewidth]{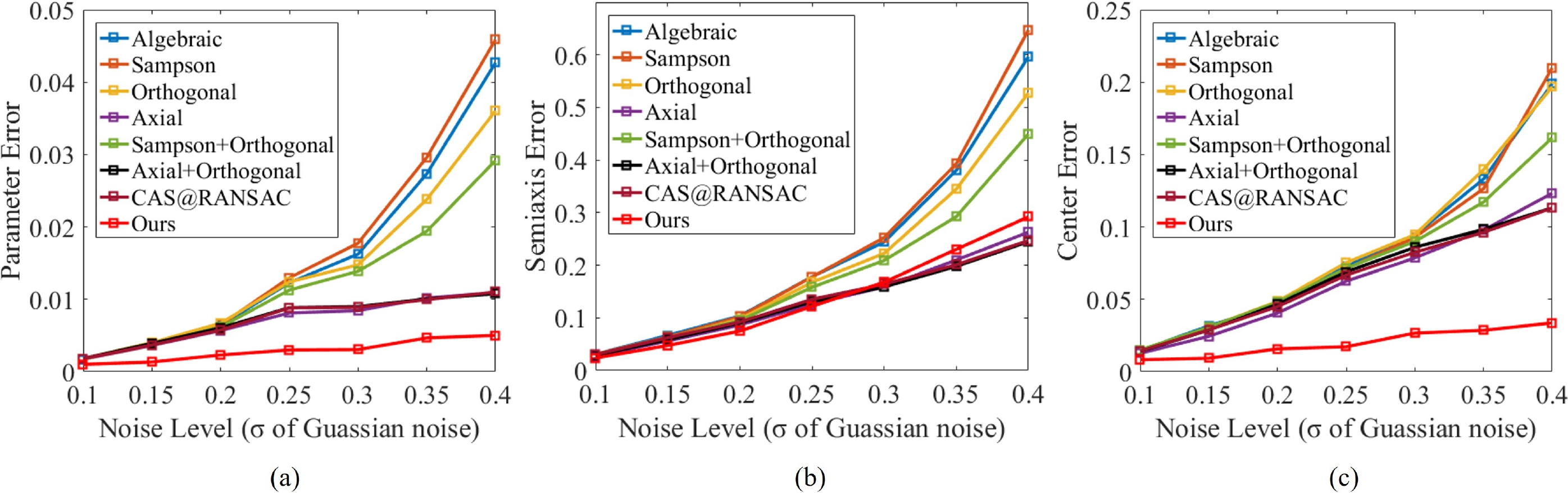}
  \caption{Fitting errors of methods using different types of distances on the Gaussian-noise dataset. (a) Parameter error. (b) Semiaxis error. (c) Center error.}
  \label{fig_9}
  \end{figure*}

1) \textbf{Accuracy on Gaussian-Noise Dataset:} Fig. 9 shows that the proposed method achieves the highest accuracy. Both the axial + orthogonal method and the CAS@RANSAC method achieve the secondarily high accuracy owing to the strict metric. Considering the complex computation of orthogonal distance, we selected CAS rather than axial + orthogonal distance to form a strict metric. Compared to CAS@RANSAC method, the proposed method achieved higher accuracy by WLS fitting using CAS. The Sampson distance and orthogonal distance are non-complementary and are combined in the Sampson + orthogonal method. Therefore, the Sampson + orthogonal method achieves the lowest accuracy among the methods of distance combination. Owing to the lack of complementarity, the methods with a single distance (except for the axial distance) perform poorly. The orthogonal method performed slightly better than the Sampson method because Sampson distance is only the first-order approximation of orthogonal distance. The proposed method achieves the highest accuracy on the entire synthetic dataset by using the strict metric and local optimization.

\begin{figure*}[!t]
  \centering
  \includegraphics[width=1 \linewidth]{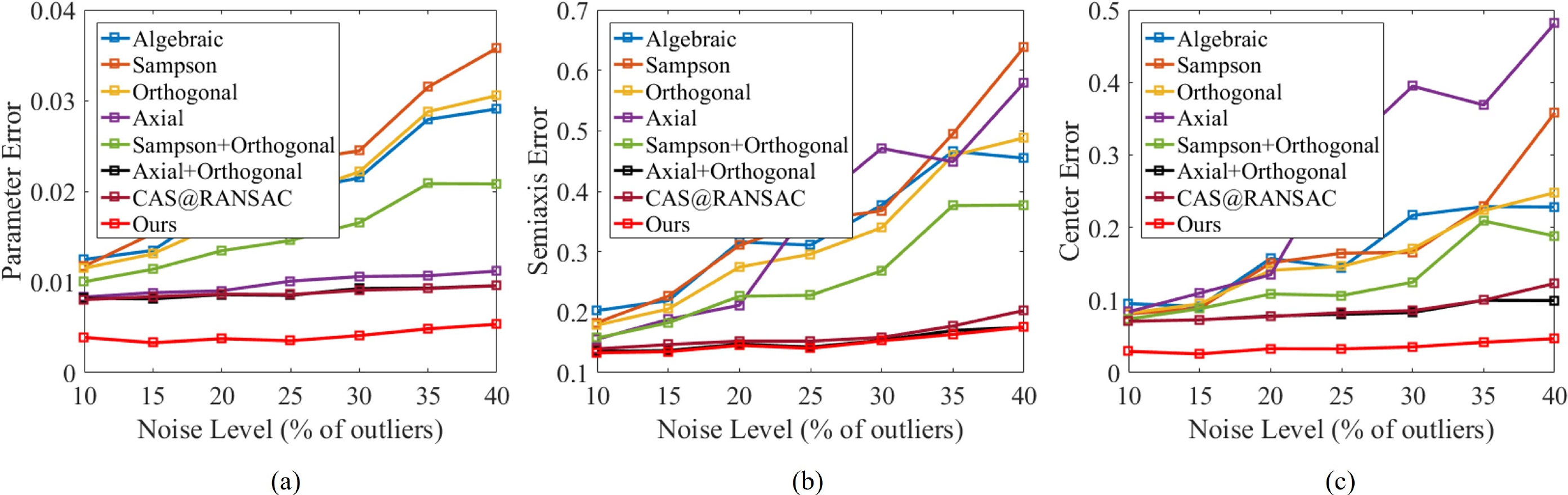}
  \caption{Fitting errors of methods using different types of distances on the outlier dataset. (a) Parameter error. (b) Semiaxis error. (c) Center error.}
  \label{fig_10}
  \end{figure*}

2) \textbf{Accuracy on Outlier Dataset:} As shown in Fig. 10, the robustness of the axial method against outliers was low in terms of semiaxis fitting and center fitting. Although the axial method achieved a high accuracy on the Gaussian-noise dataset, it achieved a low accuracy on the outlier dataset. This result indicates that the axial distance performs better than other single distances in the absence of outliers. However, the axial distance still needs to be combined with other complementary distances to provide a strict metric in the presence of outliers. The CAS@RANSAC and axial + orthogonal methods achieved higher accuracy than the Sampson + orthogonal method and single-distance-based methods. This proves that a strict metric formed by the distance combination can improve the ability to resist outliers. The results also show that a single distance failed to provide a strong constraint for the accurate model evaluation. Because the strict metric formed by CAS is applied to both the model evaluation and WLS fitting, the proposed method achieves the highest accuracy in the presence of outliers.

3) \textbf{The Required Number of Iterations:} Sample consensus searches for the best model with the highest model score in an iterative manner. According to the sample consensus, the required number of iterations is determined by (4). The required number of iterations affects fitting accuracy and algorithm speed. To further compare distance metrics, the required numbers of iterations are shown in Fig. 11.

\begin{figure}[!t]
  \centering
  \includegraphics[width=1 \linewidth]{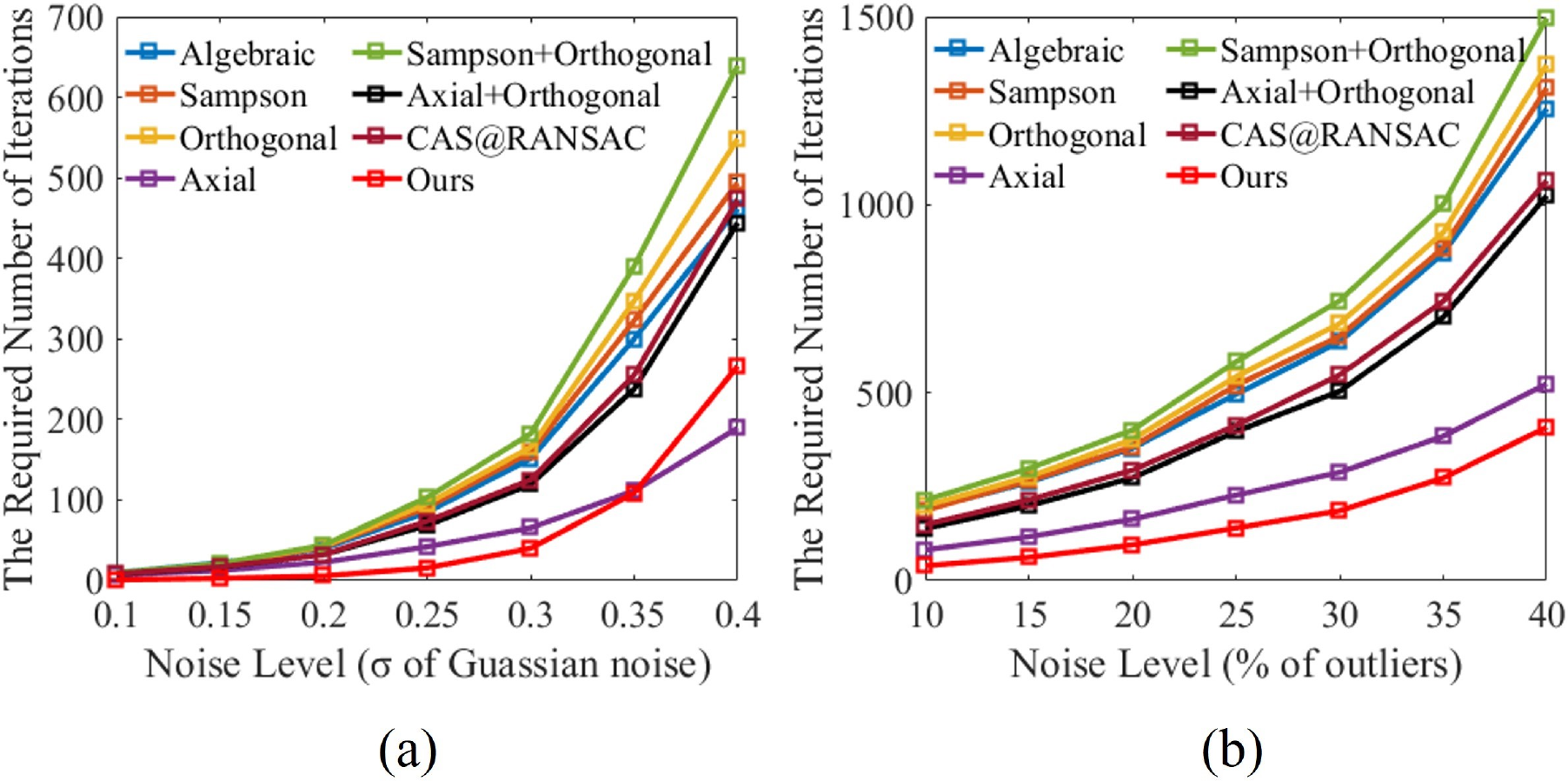}
  \caption{Required number of iterations on (a) Gaussian-noise dataset and (b) outlier dataset.}
  \label{fig_11}
  \end{figure}

The results show that the required number of iterations for the proposed method is small. Generally, the fitting accuracy of the SC-based method increases with the number of iterations and is convergent when it attains the required number of iterations \cite{A11}. The results prove that the proposed method improves the accuracy not through the manner of increasing iterations. The comparison between the proposed method and CAS@RANSAC shows that an application of local optimization reduces iterations. In local optimization, the weight of each point in WLS fitting is calculated by using the CAS. WLS fitting generates high-quality models corresponding to more inliers using the strong constraint provided by CAS. The more inliers cause less iterations according to (4). The results on two datasets show that the required number of iterations for all methods increase with the noise level because less points are classified as inliers. The required number of iterations on the outlier dataset is significantly larger than that on the Gaussian-noise dataset because a certain proportion of outliers are directly added to the outlier dataset.

\begin{figure}[!t]
  \centering
  \includegraphics[width=1 \linewidth]{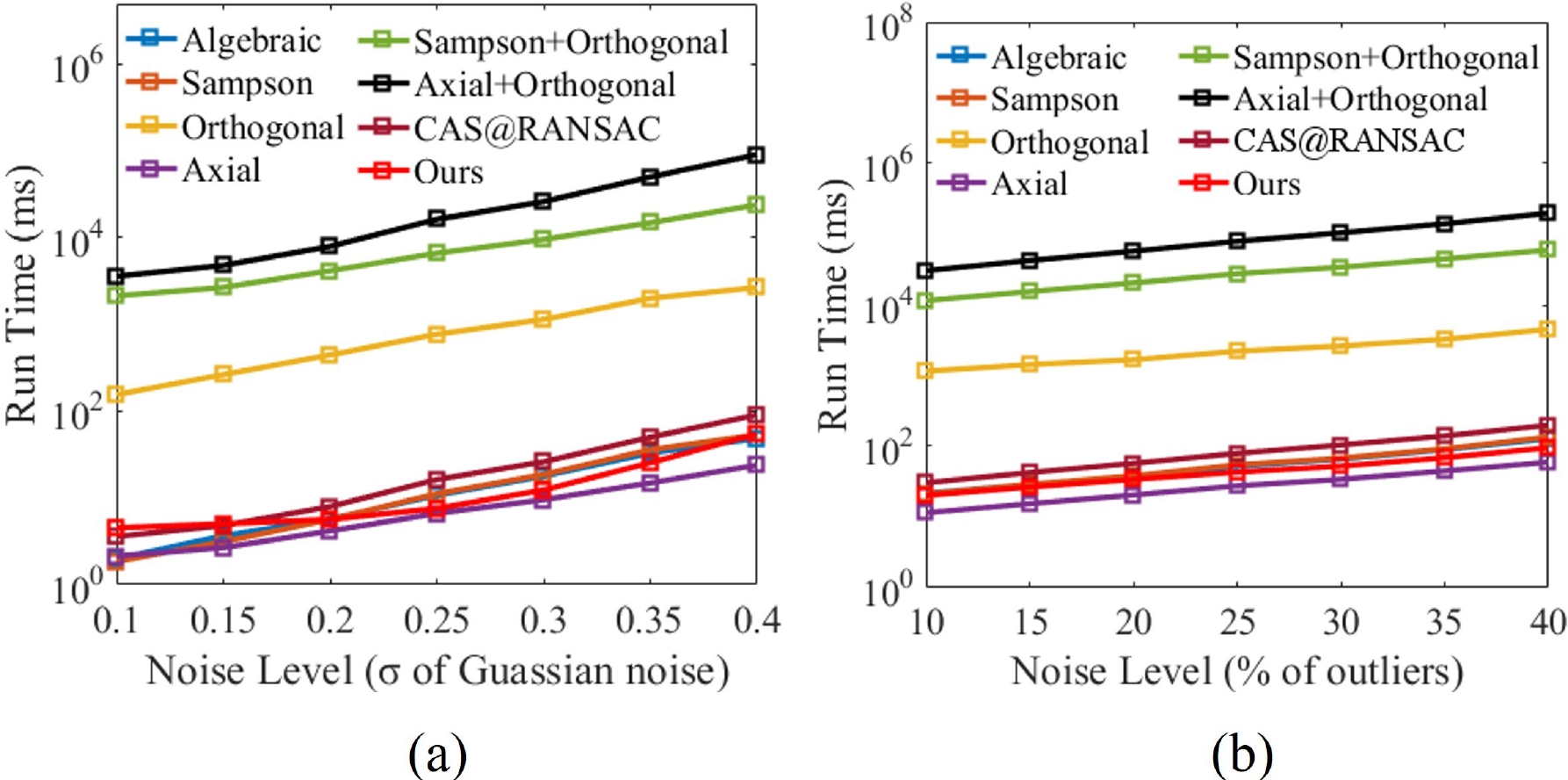}
  \caption{Algorithm speed on (a) Gaussian-noise dataset and (b) outlier dataset.}
  \label{fig_12}
  \end{figure}

4) \textbf{Algorithm Speed:} Fig. 12 shows the run time of methods using different types of distances. The Y-axis has a logarithmic scale to provide the best view. For SC-based methods, the algorithm speed was mainly determined by the required number of iterations and computational complexity of the distance metric. In conjunction with Fig. 11, it is evident that the calculated quantities of orthogonal distance are significantly larger than that of other distance metrics. This is why we selected Sampson distance rather than orthogonal distance to form a strict metric. Although the axial distance and local optimization introduce extra computations, the proposed method has less iterations. Additionally, the number of local optimizations is small because it is performed only after the best model from the RANSAC step is updated. Therefore, the proposed method achieves speed close to that of the conventional SC-based method.

We analyzed the computational complexity of an iteration to accurately evaluate the speed of the proposed method. The complexity of an iteration mainly originates from LLS fitting and the calculation of point-to-model distance. For $\mathcal{K}$-dimension model, the dimension $m$ of the parameter vector  is equal to ${{\left( {\mathcal{K} + 1} \right)\left( {\mathcal{K} + 2} \right)} \mathord{\left/
 {\vphantom {{\left( {\mathcal{K} + 1} \right)\left( {\mathcal{K} + 2} \right)} 2}} \right.
 \kern-\nulldelimiterspace} 2}$. The LLS fitting with a $n$-sized sample results in a computational complexity of $O\left( {n{m^2}} \right) + O\left( {{m^3}} \right) = O\left( {n{\mathcal{K}^4}} \right){\text{ + }}O\left( {{\mathcal{K}^6}} \right)$ [9]. To calculate the axial distance, the parameter matrix ${{\mathbf{Q}}_s}$ of a member was solved. The EVD of an upper-left $\mathcal{K} \times \mathcal{K}$ submatrix of ${{\mathbf{Q}}_1} \in {\mathbb{R}^{\left( {\mathcal{K} + 1} \right) \times \left( {\mathcal{K} + 1} \right)}}$ introduced a complexity of $O\left( {{\mathcal{K}^3}} \right)$. Assume that the input point set is denoted by $\Omega$ and the number of points in the input is $C$. Solving Sampson and axial distances introduces a polynomial calculation ${\mathbf{x}}_h^T{{\mathbf{Q}}_1}{{\mathbf{x}}_h}$, which results in computational complexity of $O\left( {C{\mathcal{K}^2}} \right)$ for the input $\Omega$. The total complexity of the proposed method in an iteration was $O\left( {n{\mathcal{K}^4} + {\mathcal{K}^6}} \right) + O\left( {{\mathcal{K}^3}} \right) + O\left( {C{\mathcal{K}^2}} \right)$, and that of RANSAC method was $O\left( {n{\mathcal{K}^4} + {\mathcal{K}^6}} \right) + O\left( {C{\mathcal{K}^2}} \right)$. The complexity of an iteration of the proposed method is the same-order as that of the RANSAC method.

\subsection{Sensitivity to Control Ratio $\lambda$ }

\begin{figure}[!t]
  \centering
  \includegraphics[width=1 \linewidth]{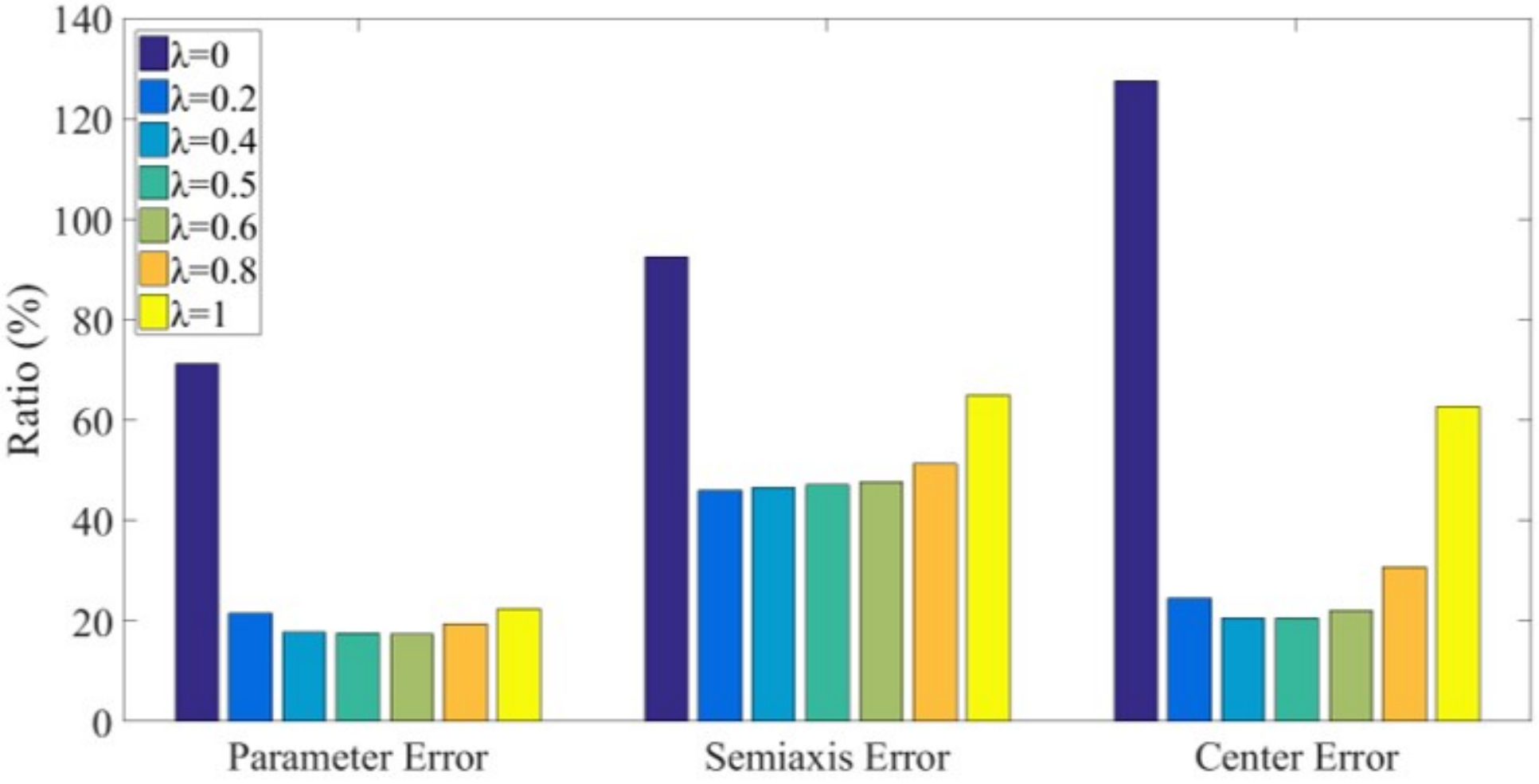}
  \caption{Sensitivity of the proposed method to control ratio $\lambda$. The ratios of errors of investigated methods to the error of Sampson method are reported.}
  \label{fig_13}
  \end{figure}

The parameter $\lambda$ in the proposed method is used to determine the weight of distance in the combination. $\lambda {\text{ = }}0.5$ denotes that the weight of axial distance is equal to that of Sampson distance in distance combination. As shown in Fig. 13, the sensitivity of the proposed method to parameter $\lambda$ is explored. For each investigated $\lambda $ value, three types of mean fitting errors on synthetic dataset were calculated. The sensitivity was evaluated using the ratios of errors of investigated methods to the error of the Sampson method.

The results show that the proposed method is insensitive to $\lambda$ when $\lambda  \in \left( {0,1} \right)$, which is an advantage for the algorithm. The dependence of CAS on the axial distance increases with $\lambda$ value. When $\lambda  = 0$ and $\lambda  = 1$, the CAS degenerates into a single distance, which is a weak metric. The results show that the fitting accuracy of $\lambda  \in \left( {0,1} \right)$ is higher than that of $\lambda  = 0$ and $\lambda  = 1$, which verifies the advantage of the combination of the complementary distances. The highest accuracy is achieved at $\lambda {\text{ = }}0.5$ because it puts the axial distance and Sampson distance on the same order of magnitude. Therefore, the set of $\lambda {\text{ = }}0.5$ is recommended in distance combination.

\subsection{Ablation Experiment}

\begin{figure}[!t]
  \centering
  \includegraphics[width=1 \linewidth]{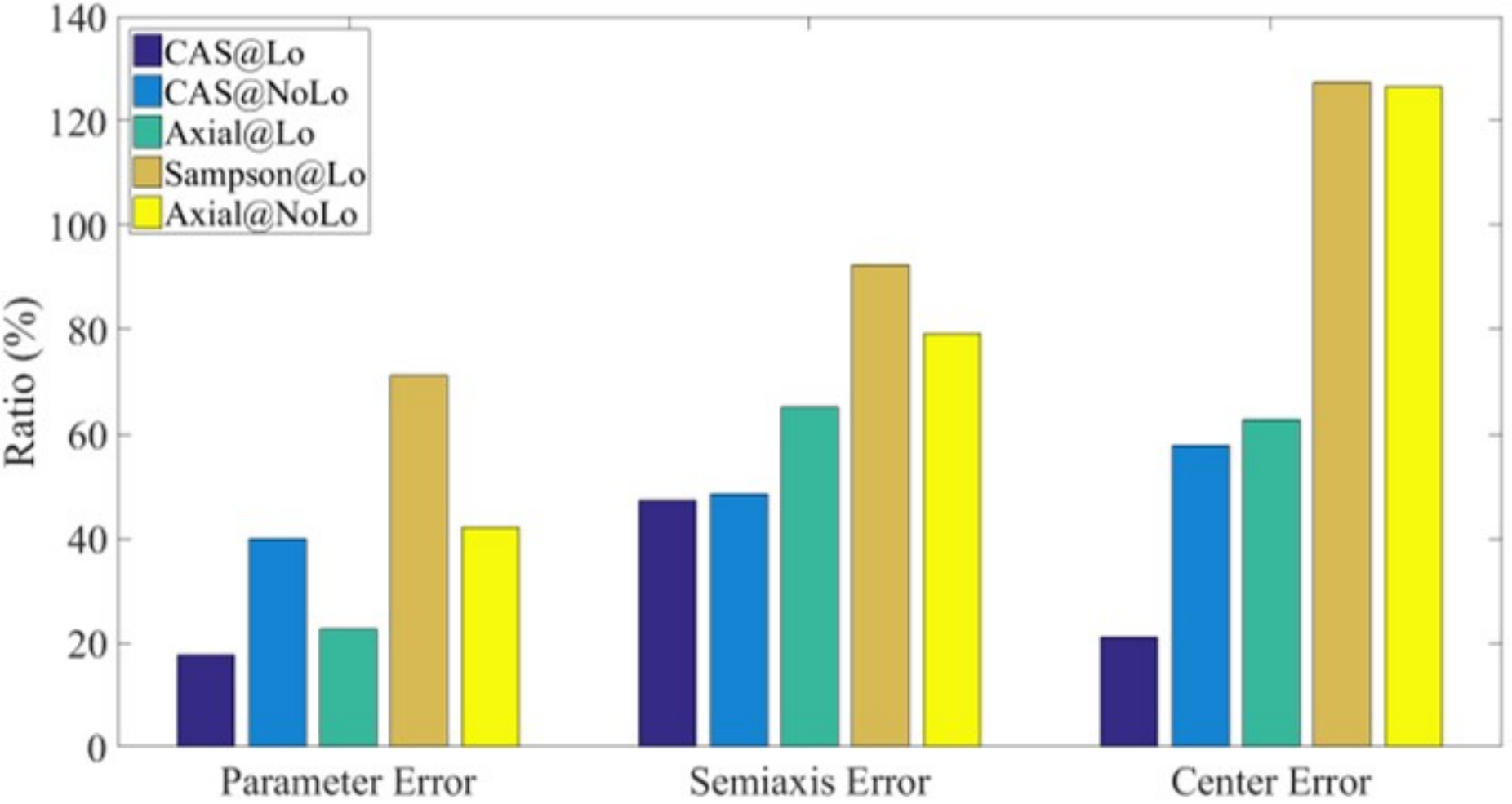}
  \caption{Ablation experiment. The ratios of errors of investigated methods to the error of Sampson method are reported. @Lo denotes the method with the local optimization, and @NoLo denotes the method without the local optimization.}
  \label{fig_14}
  \end{figure}

CAS plays a important role in two modules: mode evaluation (model score calculated using CAS) and local optimization (WLS fitting using CAS). Fig. 14 shows results of an ablation experiment to further explore the effect of each module on fitting accuracy. For each investigated method, three types of mean fitting errors on the entire synthetic dataset were calculated, and the ratios of errors of investigated methods to RANSAC were reported.
 
The results show that CAS@NoLo was more accurate than single-distance methods without local optimization, which indicates that the model evaluation using CAS was more accurate than that using a single distance. The comparison of CAS@Lo and CAS@NoLo methods indicates that local optimization can exploit CAS to further improve fitting accuracy. The ablation experiment shows that cooperative utilization of model evaluation using CAS and local optimization using CAS can significantly improve fitting accuracy.

\subsection{Experiments on Real Dataset}

The real dataset was taken from RGB-D object dataset (http://rgbd-dataset.cs.washington.edu/dataset.html) \cite{A32}. This dataset was recorded using a 3-D Kinect-style camera that can record RGB images and point clouds. The real objects that have a shape similar to an ellipsoid were selected for the real experiments. In total, 27 instances on behalf of 27 real objects were selected, and 100 runs were performed on each instance. Thus, 2700 runs were performed on the real dataset. The real objects were separated from the background. Down-sampling was used to process the raw point set to unify the number of points in input. Consequently, the input of each instance contained 500 points from the raw point set.

The ground truth of the ellipsoidal parameter is unavailable for the real objects. The SC-based methods minimize the geometrical distance whereas ADMM and SOD minimize algebraic distance. Additionally, SC-based methods are robust. For fair comparison, we only compare the robust SC-based methods by using the residuals of geometric distance metrics. The residuals of the Sampson, orthogonal, and axial distances are calculated by
\begin{eqnarray}\label{eq25}
{\text{Residual}} = \frac{{\sum\limits_{p \in \Omega } d \left( {p,\bar M} \right)}}{C},  
\end{eqnarray}

\noindent where $\bar M$ denotes the final fitting result, $C$ is the total number of points.

1) \textbf{Fitting Effect:} Fig. 15 shows the two fitting examples of the proposed method. For the real dataset, the input was only a part of the object surface. Some input points deviated from an ideal ellipsoid and outliers were common. Owing to the point classification in the sample consensus, the proposed method not only fit out an ellipsoid but also extract the inliers.

\begin{figure}[!t]
  \centering
  \includegraphics[width=1 \linewidth]{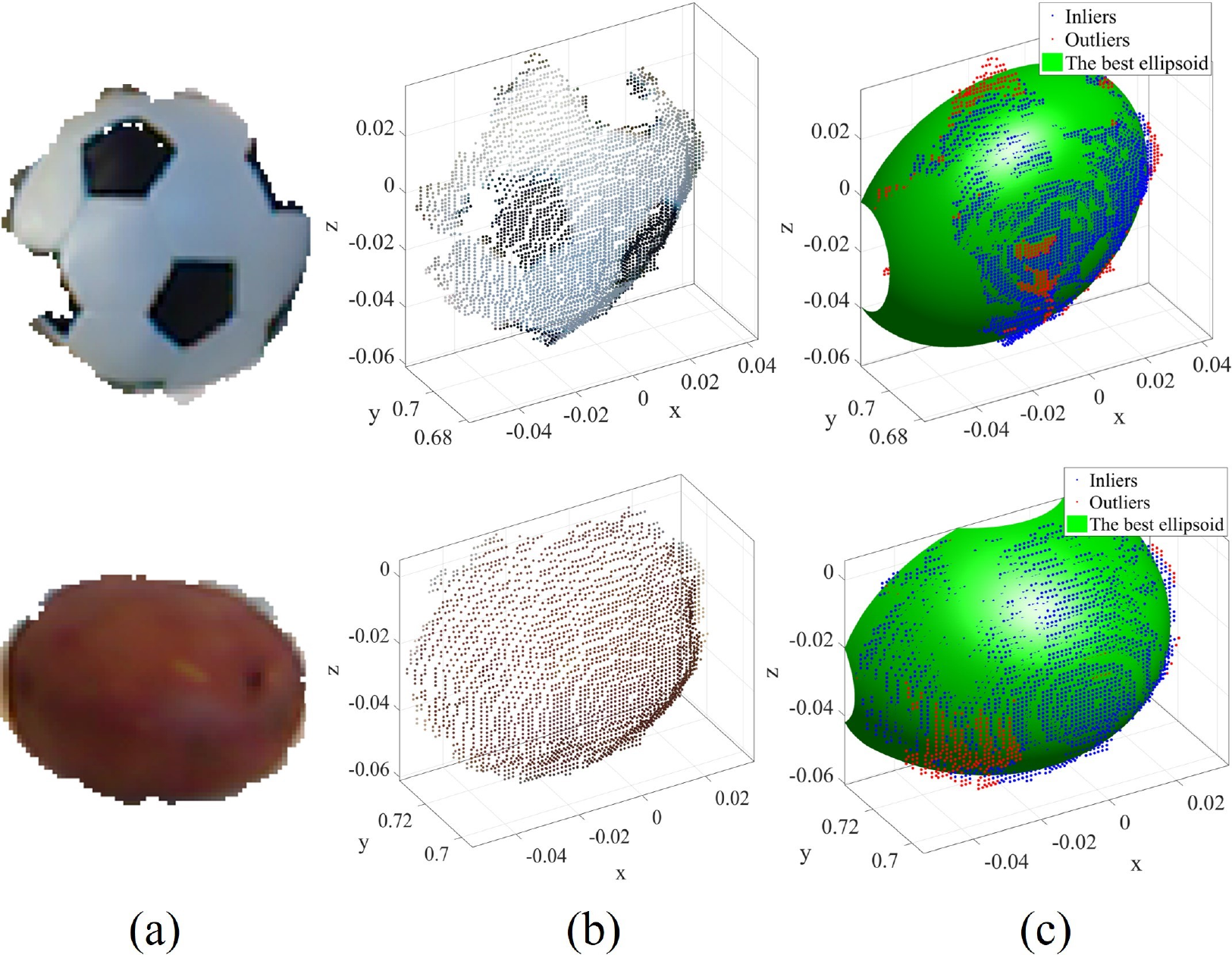}
  \caption{Fitting results of the proposed method. (a) RGB images of two instances. (b) Input point sets. (c) Fitting results.}
  \label{fig_15}
  \end{figure}

2) \textbf{Accuracy:} We only compared the accuracies of robust methods owing to the presence of outliers. Table 1 shows the mean residual on the entire real dataset. The “mean value ± standard deviation” of residual was reported. The standard deviation of accuracy was used to evaluate the stability of the algorithm. The results show the proposed method achieved the highest accuracy and stability among the SC-based methods. Although the objective of the proposed method is to minimize the combination between the axial distance and Sampson distance, it also achieved the lowest orthogonal residual.

\begin{table}[t]
  \centering
  \caption{Accuracy on Real Dataset}
  \vspace{-8 pt}
  \includegraphics[width=1 \linewidth]{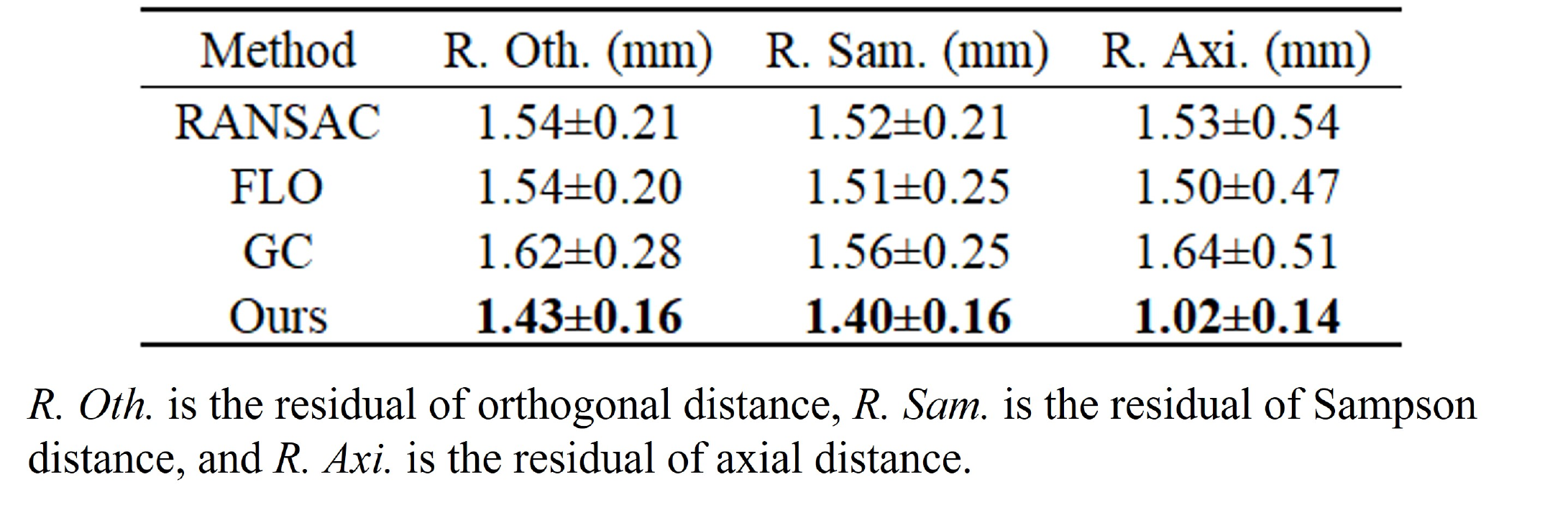}
  \label{table1}
  \vspace{-5 pt}
\end{table}

As shown in Fig. 16, the reason for a low orthogonal residual is further analyzed by the distribution of residuals on 27 instances. The Sampson residual is nearly equal to orthogonal residual because the Sampson distance is the first-order approximation of the orthogonal distance. The remarkably linear correlation indicates that the measurement ways of Sampson distance and orthogonal distance are similar. Therefore, they are non-complementary. The relation between the axial and orthogonal residuals is weak, which indicates that metrics of the axial distance and orthogonal distance are different. The combination of the complementary distances can provide a stricter metric and more constraints than a single distance for the accurate model evaluation and WLS fitting. Owing to the complementarity of distances, WLS fitting using CAS is more likely to generate high-quality ellipsoids in local optimization. Therefore, the proposed method can achieve three types of the lowest residuals.

\begin{figure}[!t]
  \centering
  \includegraphics[width=1 \linewidth]{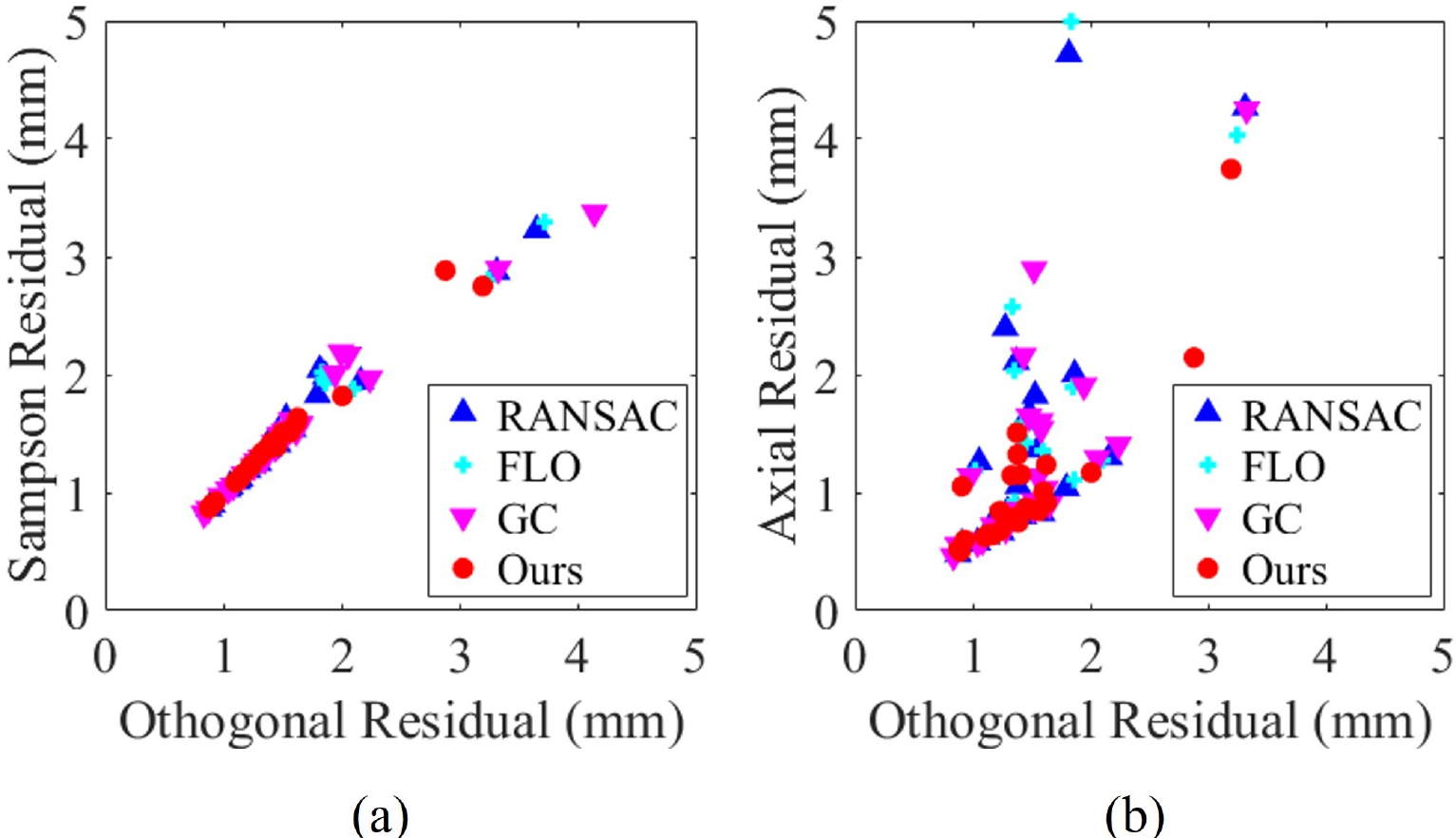}
  \caption{Distribution of residuals on 27 instances. Each scatter point denotes a residual on an instance.}
  \label{fig_16}
  \end{figure}

3) \textbf{Algorithm Speed:} Table 2 shows the mean algorithm speed on the entire real dataset. SOD and ADMM have an advantage in speed because they do not require to spend time on the point classification. For the SC-based method, the running time is related to the required number of iterations. GC and FLO also contain a local optimization. The local optimization accelerates the speed of FLO by reducing the required number of iterations. We did not show the run time of GC because it was implemented by C++ whereas other methods are implemented in MATLAB. Note that the local optimization was performed only after the best model in the RANSAC step was updated. Although the proposed method spend extra time on calculations of axial distance and local optimization, the proposed method achieved the smallest number of iterations owing to the strict metric of CAS. Therefore, the proposed method achieved a speed close to that of the SC-based methods.

\begin{table}[t]
  \centering
  \caption{Algorithm Speed and Ellipsoid Ratio on Real Dataset}
  \vspace{-8 pt}
  \includegraphics[width=1 \linewidth]{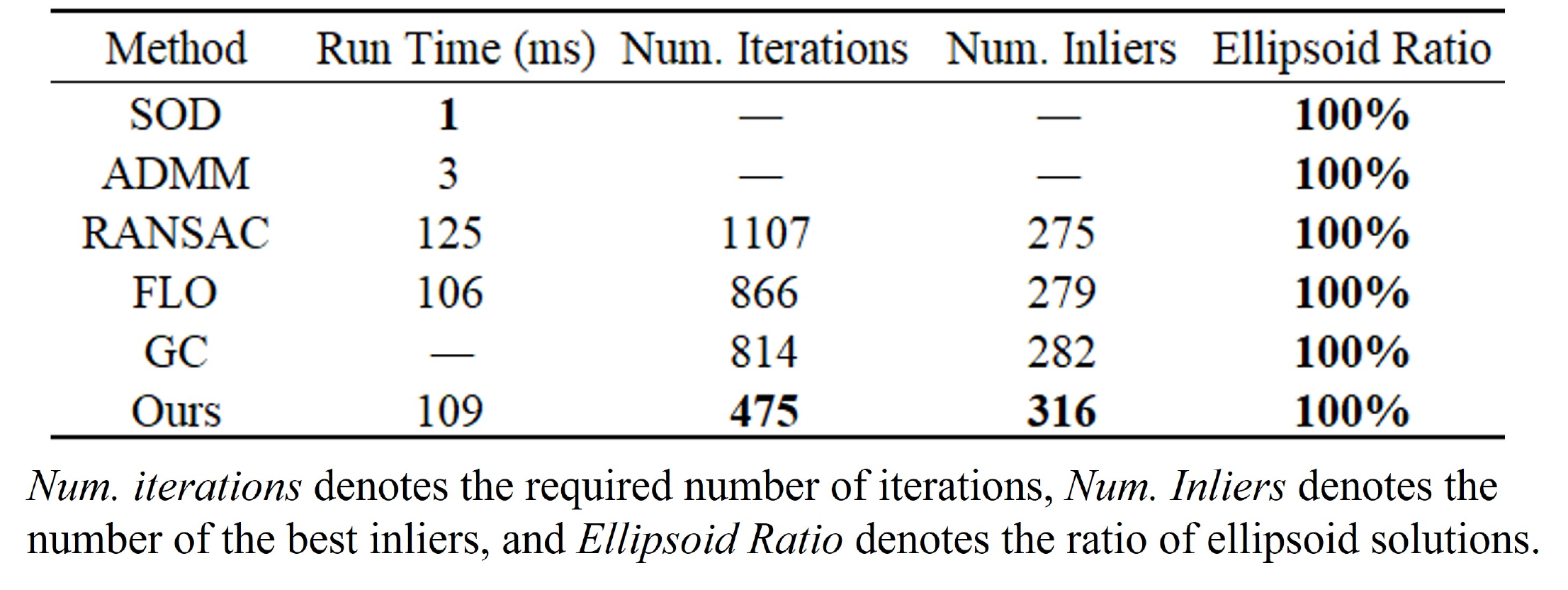}
  \label{table2}
  \vspace{-20 pt}
\end{table}

4) \textbf{Ellipsoid-Specificity:} The listed methods achieved a 100$\%$ ratio of ellipsoid solutions. The proposed method guaranteed the ellipsoid solution using the model validation. The ADMM guaranteed ellipsoid solutions without introducing extra positive semi-definiteness constraints \cite{A9}. It should be noted that the SOD has a smaller risk of non-ellipsoid solutions when its regularization item increases as reported by authors \cite{A10}.

\section{Limitation}

Although the proposed method achieved consistently high accuracy, the speed of the proposed method was slower than that of direct fitting methods. However, the speed of the proposed method was close to that of other SC-based methods, which indicates that the limitation of speed of the proposed method is mainly caused by the sample consensus. To reduce the effect of this limitation, some studies suggest more efficient sampling strategies to accelerate the generation of high-quality models \cite{A33,A34}.

\section{Conclusion}

In this study, we proposed a geometric axial distance by a notion of scaling factor to solve the nongeometric problem of algebraic distance. There exists complementarity between the axial distance and Sampson distance from the view of metric. The CAS is a strict metric and can provide more constraints than a single distance for the model evaluation and WLS fitting. Therefore, we applied CAS to point-to-model distance and obtained a novel SC-based ellipsoid fitting method. The proposed method was compared to direct fitting methods and other robust SC-based methods using different types of distances. The results showed that the proposed method has a higher robustness against outliers, consistently high accuracy, and a speed close to that of SC-based methods.
It is believed that the proposed method has significant application potential in practice owing to the presence of outliers. The cylinder can be regarded as an ellipsoid with a semiaxis of infinite length. The sphere can be regarded as an ellipsoid with three equal semiaxes. Therefore, the axial distance can also be applied to the cylinder and sphere. In the future, the proposed method will be expanded to cylinder, sphere, and hyper-ellipsoid fittings.

\ifCLASSOPTIONcompsoc
  \section*{Acknowledgments}
\else
  \section*{Acknowledgment}
\fi

This work was supported in part by the NSFC under Grant 32071680, in part by the NSFC-Xinjiang Joint Fund under Grant U1903127, and in part by the Natural Science Foundation of Shandong Province under Grant ZR2020MF052.

\ifCLASSOPTIONcaptionsoff
  \newpage
\fi

\bibliographystyle{IEEEtran}
\bibliography{ref}

\begin{thebibliography}{10}
\providecommand{\url}[1]{#1}
\csname url@samestyle\endcsname
\providecommand{\newblock}{\relax}
\providecommand{\bibinfo}[2]{#2}
\providecommand{\BIBentrySTDinterwordspacing}{\spaceskip=0pt\relax}
\providecommand{\BIBentryALTinterwordstretchfactor}{4}
\providecommand{\BIBentryALTinterwordspacing}{\spaceskip=\fontdimen2\font plus
\BIBentryALTinterwordstretchfactor\fontdimen3\font minus
  \fontdimen4\font\relax}
\providecommand{\BIBforeignlanguage}[2]{{%
\expandafter\ifx\csname l@#1\endcsname\relax
\typeout{** WARNING: IEEEtran.bst: No hyphenation pattern has been}%
\typeout{** loaded for the language `#1'. Using the pattern for}%
\typeout{** the default language instead.}%
\else
\language=\csname l@#1\endcsname
\fi
#2}}
\providecommand{\BIBdecl}{\relax}
\BIBdecl

\bibitem{A1}
G.~Calafiore, ``Approximation of n-dimensional data using spherical and
  ellipsoidal primitives,'' \emph{IEEE Trans. Syst. Man Cybern}, vol.~32,
  no.~2, pp. 269--278, 2002.

\bibitem{A2}
S.~Sivapalan, D.~Chen, S.~Denman, S.~Sridharan, and C.~Fookes, ``3d ellipsoid
  fitting for multi-view gait recognition,'' in \emph{Proc. IEEE Int. Conf.
  Adv. Video Signal-Based Surveillance}, 2011, pp. 355--360.

\bibitem{A3}
R.~Y. Da~Xu and M.~Kemp, ``Fitting multiple connected ellipses to an image
  silhouette hierarchically,'' \emph{IEEE Trans. Image Process}, vol.~19,
  no.~7, pp. 1673--1682, 2010.

\bibitem{A4}
N.~Grammalidis and M.~G. Strintzis, ``Head detection and tracking by 2-d and
  3-d ellipsoid fitting,'' in \emph{Proc. Comput. Graph. Int.}, 2000, pp.
  221--226.

\bibitem{A5}
J.~C. Fang, H.~W. Sun, J.~J. Cao, X.~Zhang, and Y.~Tao, ``A novel calibration
  method of magnetic compass based on ellipsoid fitting,'' \emph{IEEE Trans.
  Instrum. Meas.}, vol.~60, no.~6, pp. 2053--2061, 2011.

\bibitem{A6}
Z.~Y. Zhang, ``A flexible new technique for camera calibration,'' \emph{IEEE
  Trans. Pattern Anal. Mach. Intell.}, vol.~22, no.~11, pp. 1330--1334, 2000.

\bibitem{A35}
R.~S. Kothari, A.~K. Chaudhary, R.~J. Bailey, J.~B. Pelz, and G.~J. Diaz,
  ``Ellseg: An ellipse segmentation framework for robust gaze tracking,''
  \emph{IEEE Trans. Visual Comput. Graphics}, vol.~27, no.~5, pp. 2757--2767,
  2021.

\bibitem{A36}
J.~L. Liang, M.~H. Zhang, D.~Liu, and W.~Y. Wang, ``Shape fitting for the shape
  control system of silicon single crystal growth,'' \emph{IEEE Trans. Ind.
  Inf.}, vol.~11, no.~2, pp. 363--374, 2015.

\bibitem{A37}
Z.~Lu, B.~Liu, K.~Zhang, H.~Lin, and Y.~Zhang, ``A method for measuring the
  inclination of forgings based on an improved optimization algorithm for
  fitting ellipses,'' \emph{IEEE Trans. Instrum. Meas.}, vol.~72, pp. 1--11,
  2023.

\bibitem{A38}
J.~Augustyn and M.~Kampik, ``Application of ellipse fitting algorithm in
  incoherent sampling measurements of complex ratio of ac voltages,''
  \emph{IEEE Trans. Instrum. Meas.}, vol.~66, no.~6, pp. 1117--1123, 2017.

\bibitem{A39}
M.~Zakrzewski, A.~Singh, E.~Yavari, X.~M. Gao, O.~Boric-Lubecke, J.~Vanhala,
  and K.~Palovuori, ``Quadrature imbalance compensation with ellipse-fitting
  methods for microwave radar physiological sensing,'' \emph{IEEE Trans.
  Microwave Theory Tech.}, vol.~62, no.~6, pp. 1400--1408, 2014.

\bibitem{A7}
S.~J. Ahn, W.~Rauh, H.~S. Cho, and H.~J. Warnecke, ``Orthogonal distance
  fitting of implicit curves and surfaces,'' \emph{IEEE Trans. Pattern Anal.
  Mach. Intell.}, vol.~24, no.~5, pp. 620--638, 2002.

\bibitem{A8}
M.~M. Blane, Z.~B. Lei, H.~Civi, and D.~B. Cooper, ``The 3l algorithm for
  fitting implicit polynomial curves and surfaces to data,'' \emph{IEEE Trans.
  Pattern Anal. Mach. Intell.}, vol.~22, no.~3, pp. 298--313, 2000.

\bibitem{A9}
Z.~C. Lin and Y.~M. Huang, ``Fast multidimensional ellipsoid-specific fitting
  by alternating direction method of multipliers,'' \emph{IEEE Trans. Pattern
  Anal. Mach. Intell.}, vol.~38, no.~5, pp. 1021--1026, 2016.

\bibitem{A10}
M.~Kesaniemi and K.~Virtanen, ``Direct least square fitting of
  hyperellipsoids,'' \emph{IEEE Trans. Pattern Anal. Mach. Intell.}, vol.~40,
  no.~1, pp. 63--76, 2018.

\bibitem{A11}
M.~A. Fischler and R.~C. Bolles, ``Random sample consensus - a paradigm for
  model-fitting with applications to image-analysis and automated
  cartography,'' \emph{Comm. ACM}, vol.~24, no.~6, pp. 381--395, 1981.

\bibitem{A12}
C.~L. Hu, G.~Wang, K.~C. Ho, and J.~L. Liang, ``Robust ellipse fitting with
  laplacian kernel based maximum correntropy criterion,'' \emph{IEEE Trans.
  Image Process}, vol.~30, pp. 3127--3141, 2021.

\bibitem{A13}
J.~Q. Yu, H.~P. Zheng, S.~R. Kulkarni, H.~V. Poor, and Ieee, ``Outlier
  elimination for robust ellipse and ellipsoid fitting,'' in \emph{Proc. IEEE
  Int. Workshop Comput. Adv. Multi-Sensor Adaptive Process.}, 2009, pp. 33--36.

\bibitem{A14}
M.~Y. Zhao, X.~H. Jia, L.~B. Fan, Y.~Liang, and D.~M. Yan, ``Robust ellipse
  fitting using hierarchical gaussian mixture models,'' \emph{IEEE Trans. Image
  Process}, vol.~30, pp. 3828--3843, 2021.

\bibitem{A15}
R.~C. Bolles and M.~A. Fischler, ``A ransac-based approach to model fitting and
  its application to finding cylinders in range data,'' in \emph{Proc. IJCAI},
  1981, pp. 637--643.

\bibitem{A16}
F.~Q. Duan, L.~Wang, and P.~Guo, ``Ransac based ellipse detection with
  application to catadioptric camera calibration,'' in \emph{Proc. Int. Conf.
  Neural Inf. Process. Cham}, vol. 6444, 2010, pp. 525--+.

\bibitem{A17}
R.~Raguram, O.~Chum, M.~Pollefeys, J.~Matas, and J.~M. Frahm, ``Usac: A
  universal framework for random sample consensus,'' \emph{IEEE Trans. Pattern
  Anal. Mach. Intell.}, vol.~35, no.~8, pp. 2022--2038, 2013.

\bibitem{A18}
O.~Chum, J.~Matas, and J.~Kittler, ``Locally optimized ransac,'' in \emph{Proc.
  DAGM-Symp.}, vol. 2781, 2003, pp. 236--243.

\bibitem{A19}
K.~Lebeda, J.~Matas, and O.~Chum, ``Fixing the locally optimized ransac,'' in
  \emph{Proc. British Mach. Vis. Conf.}, 2012.

\bibitem{A20}
D.~Barath and J.~Matas, ``Graph-cut ransac: Local optimization on spatially
  coherent structures,'' \emph{IEEE Trans. Pattern Anal. Mach. Intell.},
  vol.~44, no.~9, pp. 4961--4974, 2022.

\bibitem{A21}
J.~L. Liang, Y.~L. Wang, and X.~J. Zeng, ``Robust ellipse fitting via
  half-quadratic and semidefinite relaxation optimization,'' \emph{IEEE Trans.
  Image Process}, vol.~24, no.~11, pp. 4276--4286, 2015.

\bibitem{A22}
G.~K. Lott, ``Direct orthogonal distance to quadratic surfaces in 3d,''
  \emph{IEEE Trans. Pattern Anal. Mach. Intell.}, vol.~36, no.~9, pp.
  1888--1892, 2014.

\bibitem{A23}
D.~Eberly, ``Distance from point to a general quadratic curve or a general
  quadric surface,'' \emph{Geometric Tools, LLC}, 2008.

\bibitem{A24}
A.~Atieg and G.~A. Watson, ``A class of methods for fitting a curve or surface
  to data by minimizing the sum of squares of orthogonal distances,'' \emph{J.
  Computational and Applied Math.}, vol. 158, no.~2, pp. 277--296, 2003.

\bibitem{A25}
P.~D. Sampson, ``Fitting conic sections to very scattered data - an iterative
  refinement of the bookstein algorithm,'' \emph{Comput. Graph. Image
  Process.}, vol.~18, no.~1, pp. 97--108, 1982.

\bibitem{A26}
Z.~L. Szpak, W.~Chojnacki, and A.~van~den Hengel, ``Guaranteed ellipse fitting
  with the sampson distance,'' in \emph{Proc. 12th Eur. Conf. Computer
  Vision.}, vol. 7576, 2012, pp. 87--100.

\bibitem{A27}
J.~L. Liang, M.~H. Zhang, D.~Liu, X.~J. Zeng, O.~Ojowu, K.~X. Zhao, Z.~Li, and
  H.~Liu, ``Robust ellipse fitting based on sparse combination of data
  points,'' \emph{IEEE Trans. Image Process}, vol.~22, no.~6, pp. 2207--2218,
  2013.

\bibitem{A28}
M.~Rouhani, A.~D. Sappa, and Ieee, ``Relaxing the 3l algorithm for an accurate
  implicit polynomial fitting,'' in \emph{Proc. IEEE Int. Conf. Comput. Vis.
  Pattern Recog.}, 2010, pp. 3066--3072.

\bibitem{A29}
Y.~H. Wu, H.~R. Wang, F.~L. Tang, and Z.~H. Wang, ``Efficient conic fitting
  with an analytical polar-n-direction geometric distance,'' \emph{Pattern
  Recognit.}, vol.~90, pp. 415--423, 2019.

\bibitem{A30}
A.~Fitzgibbon, M.~Pilu, and R.~B. Fisher, ``Direct least square fitting of
  ellipses,'' \emph{IEEE Trans. Pattern Anal. Mach. Intell.}, vol.~21, no.~5,
  pp. 476--480, 1999.

\bibitem{A31}
P.~H.~S. Torr and A.~Zisserman, ``Mlesac: A new robust estimator with
  application to estimating image geometry,'' \emph{Comput. Vis. Image
  Under-standing.}, vol.~78, no.~1, pp. 138--156, 2000.

\bibitem{A32}
K.~Lai, L.~F. Bo, X.~F. Ren, D.~Fox, and Ieee, ``A large-scale hierarchical
  multi-view rgb-d object dataset,'' in \emph{Proc. IEEE Int. Conf. Robot.
  Autom. (ICRA).}, 2011, pp. 1817--1824.

\bibitem{A33}
R.~B. Tennakoon, A.~Bab-Hadiashar, Z.~W. Cao, R.~Hoseinnezhad, and D.~Suter,
  ``Robust model fitting using higher than minimal subset sampling,''
  \emph{IEEE Trans. Pattern Anal. Mach. Intell.}, vol.~38, no.~2, pp. 350--362,
  2016.

\bibitem{A34}
O.~Chum and J.~Matas, ``Matching with prosac - progressive sample consensus,''
  in \emph{Proc. IEEE Conf. Comput. Vis. Pattern Recognit.}, 2005, pp.
  220--226.

\end{thebibliography}

\end{document}